\documentclass{article}

\PassOptionsToPackage{sort&compress,comma,square,super}{natbib}
\pdfoutput=1

\usepackage[preprint]{neurips_2023}
\usepackage[utf8]{inputenc}
\usepackage[T1]{fontenc}
\usepackage{hyperref}
\usepackage{url}
\usepackage{booktabs}
\usepackage{amsfonts}
\usepackage{nicefrac}
\usepackage{microtype}
\usepackage{xcolor}
\usepackage{enumitem}
\usepackage{graphicx}
\usepackage{amsmath}
\usepackage{natbib}
\usepackage{multibib}
\newcites{supp}{References}
\usepackage{algorithm}
\usepackage{algorithmic}
\usepackage{multirow}
\usepackage{tabulary}
\DeclareMathAlphabet      {\mathbfit}{OML}{cmm}{b}{it}

\title{Energy-Efficient Visual Search by Eye Movement and Low-Latency Spiking Neural Network}

\author{
  Yunhui Zhou \\
  School of Life Science\\
  Fudan University\\
  Shanghai, China 200433 \\
  \texttt{yhzhou17@fudan.edu.cn} \\
  \And
  Dongqi Han \\
  Microsoft Research Asia \\
  Shanghai, China 200232 \\
  \texttt{dongqihan@microsoft.com} \\
  \AND
  Yuguo Yu \\
  Research Institute of Intelligent Complex Systems \\
  Fudan University\\
  Shanghai, China 200433 \\
  \texttt{yuyuguo@fudan.edu.cn} \\
}

\begin{document}

\maketitle

\begin{abstract}
Human vision incorporates non-uniform resolution retina, efficient eye movement strategy, and spiking neural network (SNN) to balance the requirements in visual field size, visual resolution, energy cost, and inference latency. These properties have inspired interest in developing human-like computer vision. However, existing models haven't fully incorporated the three features of human vision, and their learned eye movement strategies haven't been compared with human's strategy, making the models' behavior difficult to interpret. Here, we carry out experiments to examine human visual search behaviors and establish the first SNN-based visual search model. The model combines an artificial retina with spiking feature extraction, memory, and saccade decision modules, and it employs population coding for fast and efficient saccade decisions. The model can learn either a human-like or a near-optimal fixation strategy, outperform humans in search speed and accuracy, and achieve high energy efficiency through short saccade decision latency and sparse activation. It also suggests that the human search strategy is suboptimal in terms of search speed. Our work connects modeling of vision in neuroscience and machine learning and sheds light on developing more energy-efficient computer vision algorithms.
\end{abstract}

\section{Introduction}

Biological vision is well adapted for survival in natural environment. In most cases, it needs to fulfill at least four requirements: high resolution, large visual field, low response latency, and low energy cost. However, having both high resolution and large visual field will increase the computational and energy cost, and may also increase response latency. To balance between these requirements, most vertebrates' retina has high resolution at center and low resolution in periphery \cite{Clarke1976, Rowe1976, Dunlop1987, GonzalezSoriano1995, Bozzano2000, Tyrrell2013}. This allows animals to cover a large visual field by a small number of cells, while still keeping a subjectively clear vision. For example, the monocular visual field of the human eye is $160^\circ\times135^\circ$ \cite{RH1990}, but the number of ganglion cells on retina is only 0.7--1.5 million per eye \cite{Curcio1990b}.

To compensate for the low resolution peripheral vision, many animals use the "fixation--saccade" eye movement strategy to view the world \cite{Easter1974, Collewijn1977, Paul1990, Schilstra1998, Ott2001, Land2011, Tyrrell2015, Land2019, Michaiel2020, Park2022}. To decrease response latency and energy cost, the eye movement strategy needs to be efficient and focus mostly on task-related regions. Indeed, humans can dynamically adjust fixation locations according to task demands \cite{Yarbus1967}, leading to better performance in visual search tasks than random eye movement strategy \cite{Najemnik2005}.

Biological vision also uses spiking neurons to process information. This has inspired the spiking neural network (SNN), which is in theory as computationally powerful as conventional artificial neural networks (ANN) \cite{Maass1997}. SNN running on neuromorphic chips can be much more energy-efficient than ANN running on GPU or CPU \cite{Rajendran2019} due to its event-driven computation and collocated processing and memory \cite{Schuman2022}.

In summary, \textbf{non-uniform resolution retina}, \textbf{efficient eye movement strategy}, and \textbf{computation by spiking neurons} may be crucial in supporting energy-efficient human vision. However, most current computer vision models process uniform-resolution images and is based on ANN. Their energy efficiency is unsatisfactory given that the human brain consumes only 20 Watts \cite{Kety1957}. Although a few models were trained to process non-uniform resolution images and make eye movements, their eye movement strategies had various limitations (such as requiring a fixed fixation number \cite{Ranzato2014, Mnih2014, Ba2015, Dabane2022, Alexe2012, Lyu2018, elsayed2019saccader, rangrej2023glitr}, or can only fixate at discrete set of locations \cite{Kumari2022}), and the learned strategy haven't been compared with that of human's, so it is unknown whether these models can achieve human-level performance. Besides, as far as we know, no model has fully incorporated the three properties of human vision.

\begin{figure}[htb!]
    \centering
    \includegraphics[width=0.85\textwidth]{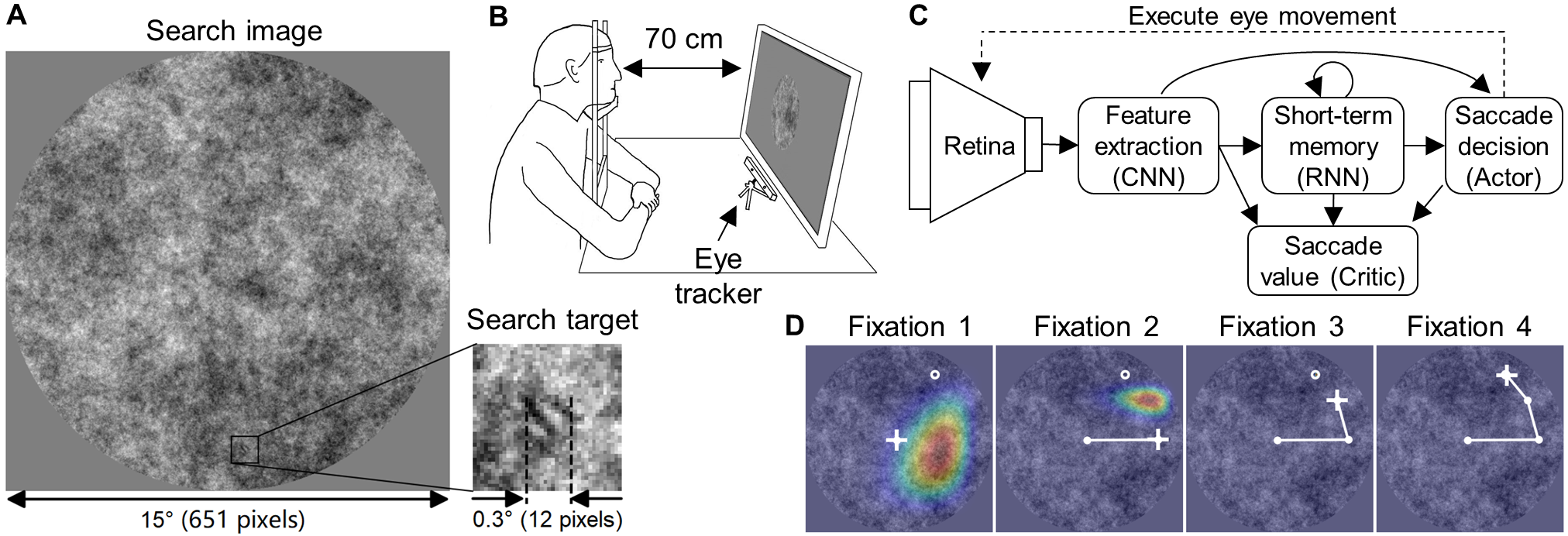}
    \caption{
        \textbf{A}. An example search image with target enlarged at lower right. Symbol "$^\circ$" denotes degrees of visual angle viewing from 70 cm distance. \textbf{B}. Human visual search experiment setup. \textbf{C}. Structure of BVSM. See Fig.~\ref{fig:model_structure} for details. \textbf{D}. An example search trial by BVSM. Cross and circle are the current fixation location and target location. Color map shows the probability distribution of the next fixation. The actual next fixation is randomly sampled from this distribution.
        }
    \label{fig:overview}
\end{figure}

The goal of this work is to establish the first bio-inspired visual search model (BVSM) that utilizes features of human vision to achieve efficient saccade decision and low energy cost (Fig.~\ref{fig:overview}C). To compare BVSM's eye movement strategy to that of human's, we train both human subjects and the model on a classical visual search task that needs to find a single Gabor target embedded in naturalistic noise image (Fig.~\ref{fig:overview}A) \citep{Najemnik2005}. The task is chosen for its following properties:
\begin{itemize}[topsep=0pt,leftmargin=0.2in]
\item \textbf{Controllable}: The features of target and background are controlled, so the target visibility across the retina can be conveniently measured by psychophysical experiment. This allows a fair comparison between BVSM's and human's search performance.
\item \textbf{Explainable}: There exists a Bayesian optimal search strategy for this task based on human target visibility data \citep{Najemnik2005, Najemnik2009}, which allows a comparison between the optimal and BVSM's search strategy. 
\item \textbf{Challenging}: The target is much smaller (1/2943) and has lower contrast than the background. Conventional network that processes full-resolution original images fails to classify whether the image contains a target.
\item \textbf{Naturalistic}: The image has $1/f$ amplitude spectrum on spatial frequencies, so it is statistically similar to natural images \cite{Burton1987} though the image contents are not the same.
\end{itemize}

BVSM can learn efficient stochastic eye movement policy by reinforcement learning (RL) to search for the target (Fig.~\ref{fig:overview}D and Supplementary Video). To summarize our contributions: 
\begin{itemize}[topsep=0pt,leftmargin=0.2in]
\item We for the first time integrate the non-uniform resolution retina, efficient eye movement strategy, and computation by SNN into BVSM. BVSM autonomously develops sparse coding during training and leverages population coding to achieve ultra-low saccade decision latency. 
\item We show that BVSM achieves high energy efficiency compared to conventional ANNs that process uniform-resolution images without making eye movements. Our work establishes a starting point for future research on human-like vision model.
\item We compare the visual search strategies of humans, BVSM, and the optimal strategy, showing that BVSM can either learn a human-like or a near-optimal strategy that outperforms humans. The comparison contributes to a deeper understanding of human eye movement strategy.

\end{itemize}

\section{Related Work}

\paragraph{Bayesian optimal visual search model}

The Bayesian optimal visual search model uses experimentally measured target visibility data to optimally compute the probability of target locations and select the next fixation location \cite{Najemnik2005, Najemnik2009} (Appendix ~\ref{elm_model_appendix}). However, this model has a drawback as its inputs are random signals generated by the target visibility data instead of the actual search image, and generating the signals requires knowing the actual target location. Therefore, the Bayesian optimal search model is not suitable for real-life applications. In contrast, the BVSM learns an efficient search strategy directly from the search image without prior knowledge of target location or target visibility data, so it is more applicable to real-life scenarios.

\paragraph{Hard attention models}

BVSM is a kind of hard attention model that down-samples the input image and makes eye movements. Currently, the most influential hard attention model architecture is probably the recurrent attention model (RAM) \citep{Mnih2014, Ba2015}. RAM contains a feature extraction network, a recurrent network (RNN), a saccade decision network, and a classification network. Other model structures usually either lack a short-term memory \citep{Dauce2020, Dabane2022}, or use simple manually designed memory \citep{Alexe2012, Ranzato2014, Akbas2017, elsayed2019saccader}. Some models also use Transformer to integrate information from multiple fixations \citep{rangrej2022consistenc, rangrej2023glitr}. BVSM has a similar architecture to RAM, but it is implemented as an SNN. We do not use a Transformer architecture because humans may only have a memory capacity of fewer than 20 items in visual search \citep{McCarley2003, Beck2006, Dickinson2007, Kaunitz2016}. Conventional RNNs should be able to handle a memory of this size.

In terms of the eye movement behavior, some previous models can only fixate at a pre-defined set of discrete locations \citep{rangrej2021probabilistic, rangrej2022consistenc, Kumari2022}, or require setting a fixed number of fixations for each trial \citep{Alexe2012, Mnih2014, Ranzato2014, Ba2015, Dabane2022}. In particular, RAM-based model may perform worse when fixating more times than it was trained on \citep{elsayed2019saccader}. BVSM makes continuous-valued saccade decisions, and it naturally employs a variable number of fixations to do visual search by generating a confidence rating to predicted target location. 

Many previous works are tested on visual search for MNIST-style targets, and the tasks are relatively easy because the search target is large ($>$4.79\%) compared to the background image \citep{Ranzato2014, Mnih2014, Ba2015, Dauce2020, Dabane2022, Kumari2022}. We find that the same issue also exists for models that were tested on natural images \citep{Alexe2012, Alexe2012} by inspecting the datasets used for training (PASCAL VOC 2007/2010 \citep{pascal-voc-2007, pascal-voc-2010}). Training hard attention models to search for tiny objects has not been thoroughly studied yet. BVSM is tested on a challenging visual search task that has been studied in neuroscience \citep{Najemnik2005, Najemnik2008, Najemnik2009, Zhou2021}. The search target is only 0.034\% of the size of the search image, and humans can only achieve a mean correct response rate of about 86\% \citep{Zhou2021}. It is interesting to see how hard attention models will perform in this task. 

In summary, various limitations exist in previous hard attention models, which we aim to overcome in our model. Table~\ref{table:compare_previous_work} shows a detailed comparison between BVSM and previous works.

\paragraph{Population coding with SNN}

The SNN's output variables are often encoded by the firing rate of output neurons. To calculate the firing rate, SNN needs to process the input for a number of time steps, which is called the inference time or latency. Lower latency means faster inference and lower energy cost, but it also limits the information carried by the firing rate of a single neuron. Previous works on using SNN in regression task (like objection detection) usually have high latency (100-3,500 time steps) to precisely calculate the firing rate \citep{Kim2020, Kim2021, Chakraborty2021, Miquel2021, Shen2022, Patel2019}, but this decreases the energy efficiency of SNN. To decrease the latency while maintaining the regression accuracy, we can use the weighted sum of the firing rate of a neuron population to encode a variable. Recently, SNN with population coding has achieved good performance on a variety of continuous-action RL tasks, while lowering the latency to only 5 time steps \citep{Tang2020, Xu2022pop}. In this work, we use population coding to achieve very low saccade decision latency (4 time steps). Table~\ref{table:compare_snn_latency} shows a comparison of our SNN to previous models.

\section{Methods}

\hypertarget{visual-search-task}{%
\subsection{Visual Search Task}\label{visual-search-task}}

The visual search task is to find a single Gabor target embedded in a circular naturalistic noise image (\(1/f\) amplitude spectrum) (Fig.~\ref{fig:overview}A) \citep{Najemnik2005, Zhou2021}. The background noise image has a diameter of 15$^\circ$ (651 pixels) with root-mean-squared (standard deviation of pixels divided by the mean) contrast of 0.2. The target is a 6 cycles/degree Gabor with a diameter of 0.3$^\circ$ (12 pixels), oriented 45$^\circ$ counterclockwise from the vertical, and windowed by a symmetrical raised cosine. The screen to display the image has a physical size of $54.3744 \times 30.2616$ cm, resolution of $1920 \times 1080$, and was placed 70 cm away from human subjects (Fig.~\ref{fig:overview}B). We used the same target contrast range in human experiment and modeling (detailed below). The background noise is varied in each trial.

\subsection{Human Experiment}\label{human-experiment}

29 human subjects (14 females, aged 18-30) participated in the experiment. For 15 subjects, we first used a two-interval force-choice (2IFC) task to measure the target visibility $d'$ at fovea as a function of target contrast, and individually estimated the target contrast for $d'=3.0$. These target contrasts (ranged 0.11-0.136) were used for all following experiments of each subject. We then conducted another 2IFC task to measure target visibility at various peripheral locations. Finally, subjects finished 200-400 visual search trials (Fig.~\ref{fig:overview}B). Visual search always started from fixating at image center. The rest 14 subjects only participated in the visual search experiment with a fixed target contrast of 0.15. We call the first 15 subjects "Low TC (target contrast) Group", and the rest 14 subjects "High TC Group". See Appendix \ref{human-experiment-details} for the details of the detection and visual search experiment.

\subsection{Bio-inspired Visual Search Model}

Fig.~\ref{fig:overview}C and Fig.~\ref{fig:model_structure} show the structure of BVSM. It contains an artificial retina and four networks.

\textbf{Artificial retina.}
We use Foveal Cartesian Geometry (FCG) \citep{Martinez2006} to non-uniformly down-sample the original image (Fig.~\ref{fig:model_structure}A). FCG copies pixels inside the fovea, and samples pixels in periphery from a series of rings with increasingly larger widths (Appendix~\ref{artificial-retina-appendix}). The sampled pixels are then assembled to form a square image as the output. We set the size of the output image to be 224×224 with a fovea size of 16×16 pixels. To restrict the entire search image within the visual field, we sample from a region twice the diameter of the search image. The configurations of artificial retina are closely related to visual search performance. In Appendix \ref{cnn-detection-appendix}, we also examined the model's target visibility at peripheral locations using a similar 2IFC task as in human experiment.

\textbf{Feature extraction network (FEN).}
The FEN processes retinal-transformed images by 7 convolution blocks and 3 linear heads (Fig.~\ref{fig:model_structure}B), and outputs the current fixation location (2D vector), target location relative to the fixation location (2D vector), and the estimated error between predicted and true target location (scalar). The output locations are all in the original full-resolution image's coordinate space. FEN uses QCFS function \cite{bu2021optimal} as activation function: $f\left( x \right) = \lambda \cdot \text{clip}\left( \frac{1}{T} \text{floor}(\frac{xT}{\lambda} + 0.5),0,1 \right)$, where \(\lambda\) is a trainable parameter and \(T = 4\). In the last layer of fully connected heads, we linearly transform the output of a population of neurons into the output vectors (population coding). FEN is converted to SNN with latency $T=4$ after training (Appendix \ref{cnn-structure-appendix}).

\textbf{Recurrent network (RNN).}
We propose a novel spiking RNN that uses 64 integrate-and-fire (IF) neurons (equation \ref{eq:IF_neuron}) to memorize fixation locations (Fig.~\ref{fig:model_structure}C). The network's inputs are the predicted fixation location from FEN and the RNN hidden state. Spiking RNN adds another temporal dimension (the SNN inference time in each fixation) to the existing temporal dimension (sequence of fixations). To maintain the real-time processing property of SNN, we set the RNN hidden state as the output spikes from the last time step of the previous fixation, and the hidden state is used in computation at the first time step of the next fixation. See Appendix \ref{spiking-rnn-appendix} and Algorithm \ref{alg:SRNN} for details.

\textbf{Actor and critic networks.}
The actor network makes saccade decisions (Fig.~\ref{fig:model_structure}D). If the estimated target location error from FEN is lower than a threshold (0.58$^\circ$, or 25 pixels), the next fixation location will be sampled near the predicted target location; otherwise the actor will process the RNN output by an SNN which outputs the mean and covariance of the next fixation location's distribution by population coding. The next fixation location can be anywhere within the square region enclosing the circular search image. The critic network measures saccade value (Fig.~\ref{fig:model_structure}E). If the estimated target location error is low, it will calculate the Q-value by the next fixation location and the predicted target location; otherwise it will calculate the Q-value by the next fixation location and RNN output. The critic network is not a spiking network, as it is not used in inference after training. See Appendix \ref{actor-critic-appendix} for details of the actor and critic networks.

\subsection{Reward Functions}\label{reward-function}

The agent receives the sum of two negative rewards after each saccade. The first reward is inhibition of return (IOR) that tracks the location of the most recent 8 fixations based on previous human research \cite{Beck2006, Kaunitz2016}, and discourages the agent to fixate at previously fixated regions (Fig.~\ref{fig:reward_functions}A,B). The IOR reward after a saccade to the \(\left( f + 1 \right)^{\text{th}}\) fixation location is a semi-circle function centered at each previously fixated location:
\begin{align} \label{eq:ior_reward}
R_{\text{IOR}} = \min_{i \in \left\{ 0\cdots7 \right\}}\left( - \frac{1}{r}\sqrt{\max\left( r^{2} - d_{f + 1,f - i}^{2},0 \right)} \right),
\end{align}
where \(d_{f + 1,f - i}\) is the distance between the \(\left( f + 1 \right)^{\text{th}}\) and \(\left( f - i \right)^{\text{th}}\) fixation, and \(r\) is the radius of the semi-circle function. 

The second reward is the saccade amplitude reward that discourages making large saccades (Fig.~\ref{fig:reward_functions}C,D). Making large saccade costs more time and energy \cite{Baloh1975}, and induces strong motion blur that disrupts visual processing \cite{Bridgeman1975}. We test two different saccade amplitude reward functions:
\begin{gather} 
R_{\text{SA}} = -0.5 + 0.5\times[\exp{(-\text{Saccade\ amplitude}/2.5^\circ)}-1], \label{eq:sacamp_reward_exp} \\
R_{\text{SA}} = -\text{Saccade\ amplitude}/7.5^\circ. \label{eq:sacamp_reward_linear} 
\end{gather}
To learn human-like search strategy, we set \(r = 0.5{^\circ}\) for IOR reward, and use equation \ref{eq:sacamp_reward_exp} for saccade amplitude reward. We call this setting "HP (hyperparameter) 1". To learn Bayesian-inference-like search strategy, we set \(r = 2.5{^\circ}\) for IOR reward, and use equation \ref{eq:sacamp_reward_linear} for saccade amplitude reward. We call this setting "HP 2". We also test BVSM's performance under some changes to the two reward functions (Table~\ref{table:hyperparameter_scan}). We set the discount factor \(\gamma = 0.95\).

\subsection{BVSM Training and Evaluation}\label{model-training-evaluation}

BVSM is trained in three stages. First, we train the FEN to detect target in random samples of retinal transformed images. Second, we convert the FEN to SNN, and fine-tune the SNN. Third, we fix the FEN, and use soft actor-critic (SAC) \citep{Haarnoja2018} to train the RNN, actor, and critic to do visual search. We directly train the SNN in RNN and actor by spatio-temporal backpropagation (STBP) \citep{Yujie2018}. Table~\ref{table:hyperparameter_list} shows all hyperparameters. Fig.~\ref{fig:cnn_training_process}, \ref{fig:rl_training_process_hp1}, and \ref{fig:rl_training_process_hp2} show the loss and task performance during each training stage. BVSM is evaluated for 10,000 trials after training. Details are shown in Appendix \ref{model-training-appendix}. 

\subsection{Automatic Search Termination}\label{automatic-search-termination}

Search trial ends either by reaching the maximum fixation number (50 in training, 200 in testing), or by a ``double check and stop'' rule. The latter is triggered when the estimated target location error of two consecutive fixations is below 0.58$^\circ$ (25 pixels) and the distance between the two predicted target locations is less than 0.5$^\circ$. If any of the final two fixations are within 1$^\circ$ away from the true target location (same as in human experiment), the trial is labeled as correct; otherwise error. If a trial does not automatically terminate when reaching the maximum fixation number, it is labeled as error.

\section{Results}

\subsection{Target Detection Performance Comparison}\label{detection-performance}

\begin{figure}[htb!]
    \centering
    \includegraphics[width=0.8\textwidth]{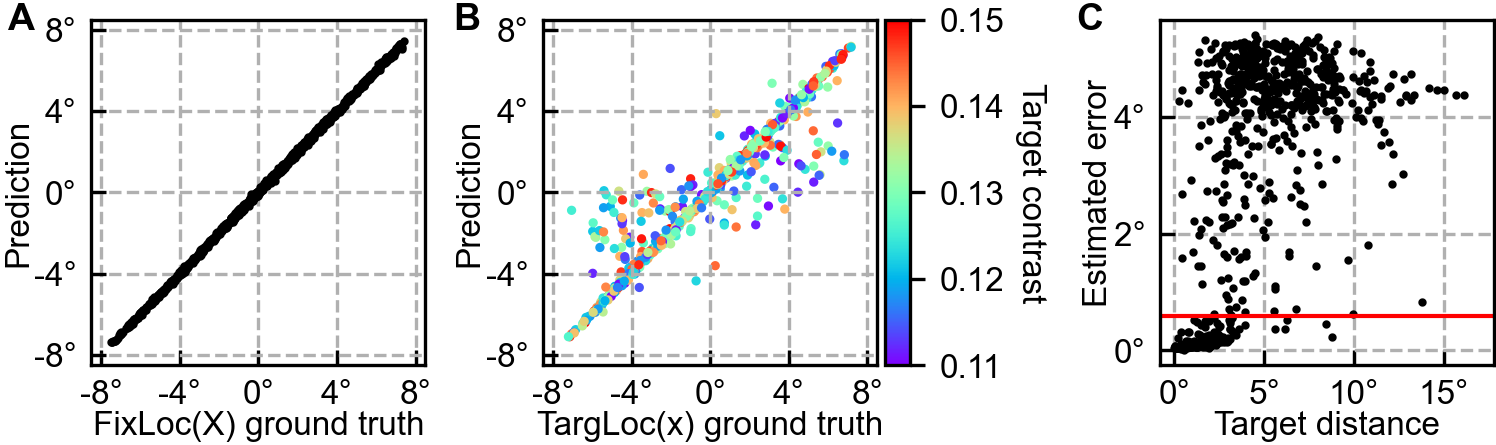}
            \caption{FEN regression results of 1024 random retinal transformed images. \textbf{A}: Relationship between predicted and true fixation location. 0$^\circ$ represents image center. \textbf{B}: Relationship between predicted and true target location within 4.0$^\circ$ away from fixation location. A and B only show the regression results for horizontal location. The results for vertical location are similar. \textbf{C}: Relationship between target distance and estimated target location error. The red horizontal line shows the decision threshold (0.58$^\circ$, 25 pixels) used in visual search.
        }
    \label{fig:cnn_regression_results}
\end{figure}

The spiking FEN accurately regresses the current fixation location (Fig.~\ref{fig:cnn_regression_results}A) and target location when the distance between target and fixation location is small and the target contrast is high (Fig.~\ref{fig:cnn_regression_results}B). FEN outputs a low estimated target location error when the target locates near the fixation, which means a high prediction confidence; otherwise it outputs a high estimated error. We can set the target detection threshold as 25 pixels (Fig.~\ref{fig:cnn_regression_results}C). If the estimated error is below the threshold, the network will fixate at the predicted target location in the next fixation. 

\begin{figure}[htb!]
    \centering
    \includegraphics[width=0.9\textwidth]{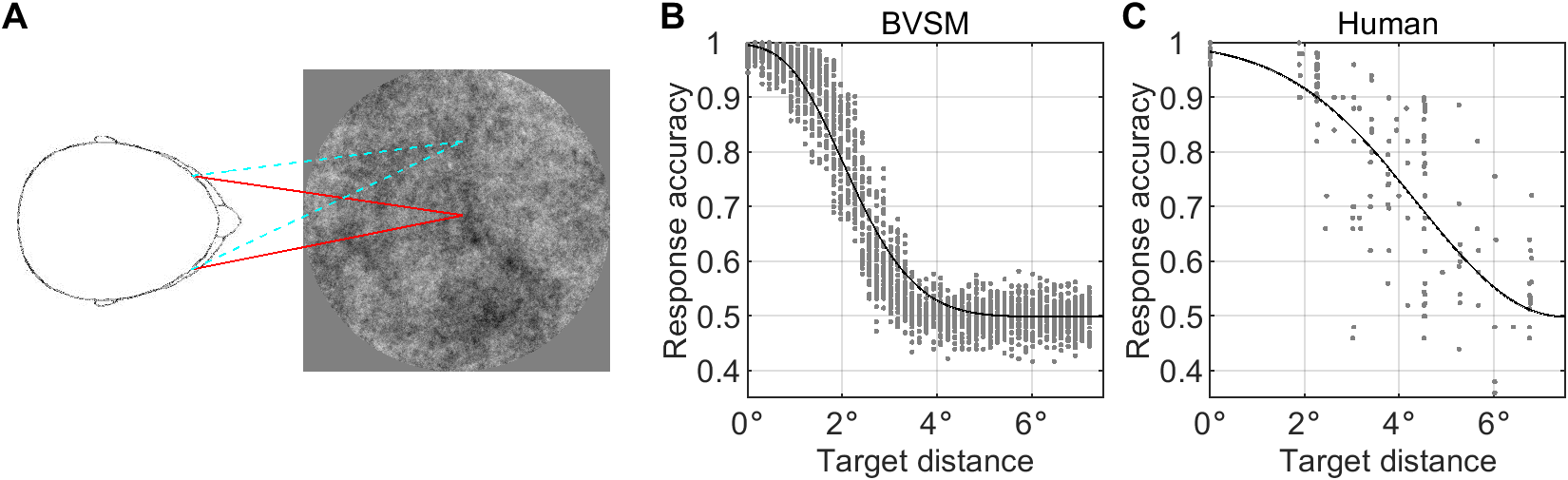}
    \caption{Detection task performance of BVSM and humans. \textbf{A}: In detection task, subjects are required to keep fixating at image center (red solid line), but report whether target is present at a peripheral location (blue dashed line) within a single fixation. \textbf{B}, \textbf{C}: Relationship between target eccentricity and response accuracy. In B, each gray dot represents a combination of target eccentricity, target angular position, and target contrast. In C, each gray dot represents a combination of subject index, target eccentricity, and target angular position. The black curve represents the fitted Weibull function (with inverted x-axis).}
    \label{fig:detection_fixnum}
\end{figure}

In the detection task (Fig.~\ref{fig:detection_fixnum}A), FEN performs similar to humans in central vision, but worse than humans in peripheral vision (Fig.~\ref{fig:detection_fixnum}B,C). In addition, conventional ANN that processes uniform-resolution images fails to detect the small target (Appendix \ref{full_resolution_model_appendix}). This shows the importance of using non-uniform resolution retina when needing to discriminate fine details from a large background.

\begin{figure}[htb!]
    \centering
    \includegraphics[width=0.85\textwidth]{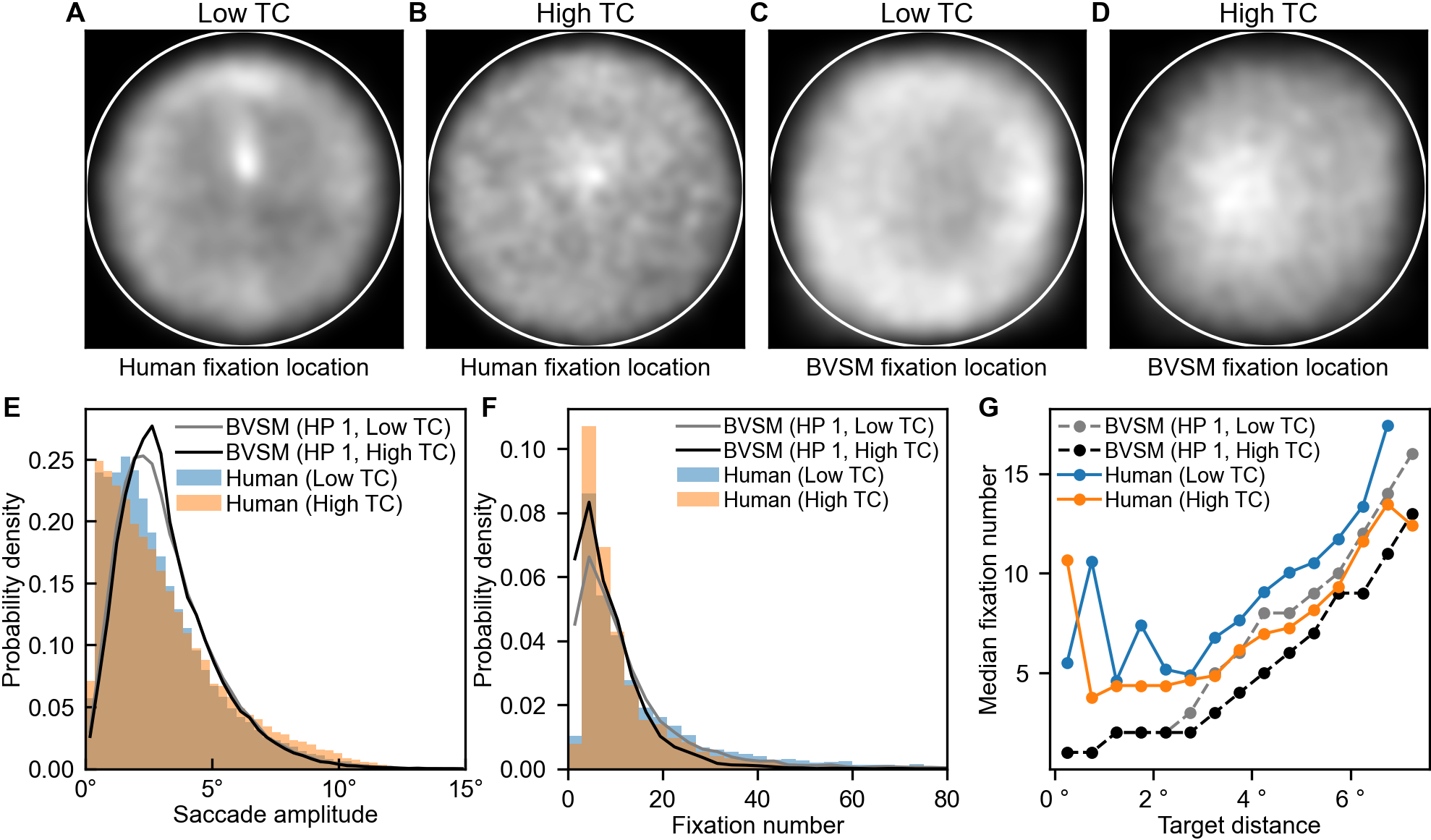}
    \caption{BVSM and human eye movement metrics. \textbf{A-D}: Distribution of fixation location. TC: target contrast. \textbf{E}. Distribution of saccade amplitude. \textbf{F}: Distribution of fixation number in correct visual search trials. \textbf{G}: Median fixation number as a function of target eccentricity in correct trials. In this figure, BVSM is trained under hyperparameter group 1 (HP 1).} 
    \label{fig:model_human_em_metrics}
\end{figure}

\subsection{Visual Search Strategy Comparison}\label{eye-movement-metrics}

After training with hyperparamter group 1 (HP 1), BVSM's search strategy shows some similarity with humans' strategy. The distribution of fixation location of both BVSM and humans has weak doughnut shape when searching for low contrast target (Fig.~\ref{fig:model_human_em_metrics}A,C), and does not show such feature when searching for high contrast target (Fig.~\ref{fig:model_human_em_metrics}B,D). 

BVSM and humans exhibit similar saccade amplitude (Fig.~\ref{fig:model_human_em_metrics}E) and fixation number distributions (Fig.~\ref{fig:model_human_em_metrics}F). Both of them favor small saccades. BVSM more frequently completes search with very few fixations, especially when the target is near the initial fixation location (Fig.~\ref{fig:model_human_em_metrics}G). BVSM quickly finds the target in such cases, while humans require more fixations. As the target distance increases, the search speeds of BVSM and humans become closer (Fig.~\ref{fig:model_human_em_metrics}G).

In addition, BVSM has higher response accuracy and search speed than humans for both low and high contrast targets (Table~\ref{table:search_performance}). In particular, it surpasses human response accuracy by about 11\% for low contrast target. The model is better at searching than humans despite having lower target visibility in peripheral vision.

Although BVSM searches better, its search strategy is limited to circular scanning (Fig.~\ref{fig:model_human_scanpath}A). While humans may use similar scanpath, they exhibit a more diverse search strategy, such as repeatedly fixating on suspected target locations (Fig.~\ref{fig:model_human_scanpath}B, blue) and scanning the search image even when the target is close to the initial fixation location (Fig.~\ref{fig:model_human_scanpath}B, yellow). This may explain why humans search more slowly when the target distance is small. Additionally, humans sometimes use irregular scanpaths with short saccades (Fig.~\ref{fig:model_human_scanpath}B, green), but the reason for this is currently unclear.

\begin{table}[htb!]
\caption{Search performance comparison. BVSM results are averaged from models trained with 5 random seeds. Human results are averaged from all subjects. ELM results are averaged from 5 independent evaluations, each consisting of 10,000 trials. Mean and median fixation number (FixNum) are calculated from correct trials. For the ELM model, target visibility ($d'$) of 3.0 corresponds to target contrast of 0.11-0.136 (Appendix \ref{supp_detection_exp_procedure}). HP 1 and HP 2 means hyperparameter gourp 1 and group 2. Performance data are shown in mean $\pm$ standard deviation.}
\begin{center}
\begin{tabular}{ccccc}
\toprule
             & Target contrast & Percent correct   & Med(FixNum)  & Mean(FixNum) \\ \midrule
BVSM (HP 1)  & 0.11--0.136     & $97.8 \pm 0.2\%$  & $8.4 \pm 0.5$  & $13.3 \pm 0.3$      \\ 
BVSM (HP 1)  & 0.15            & $99.5 \pm 0.1\%$  & $6.0 \pm 0.0$  & $8.4 \pm 0.2$       \\ 
Human (n=15) & 0.11--0.136     & $86.5 \pm 6.3\%$  & $10.0 \pm 2.7$ & $16.7 \pm 4.7$      \\
Human (n=14) & 0.15            & $97.5 \pm 1.5\%$  & $8.0 \pm 1.9$  & $13.4 \pm 4.5$      \\ \midrule
BVSM (HP 2)  & 0.11--0.136     & $98.2 \pm 0.1\%$  & $7.0 \pm 0.0$  & $10.8 \pm 0.1$ \\ 
ELM          & $d'=3.0$        & $97.8 \pm 0.0\%$  & $8.0 \pm 0.0$  & $8.8 \pm 0.1$  \\ \bottomrule
\end{tabular}
\end{center}
\label{table:search_performance}
\end{table}

\begin{figure}[htb!]
    \centering
    \includegraphics[width=0.8\textwidth]{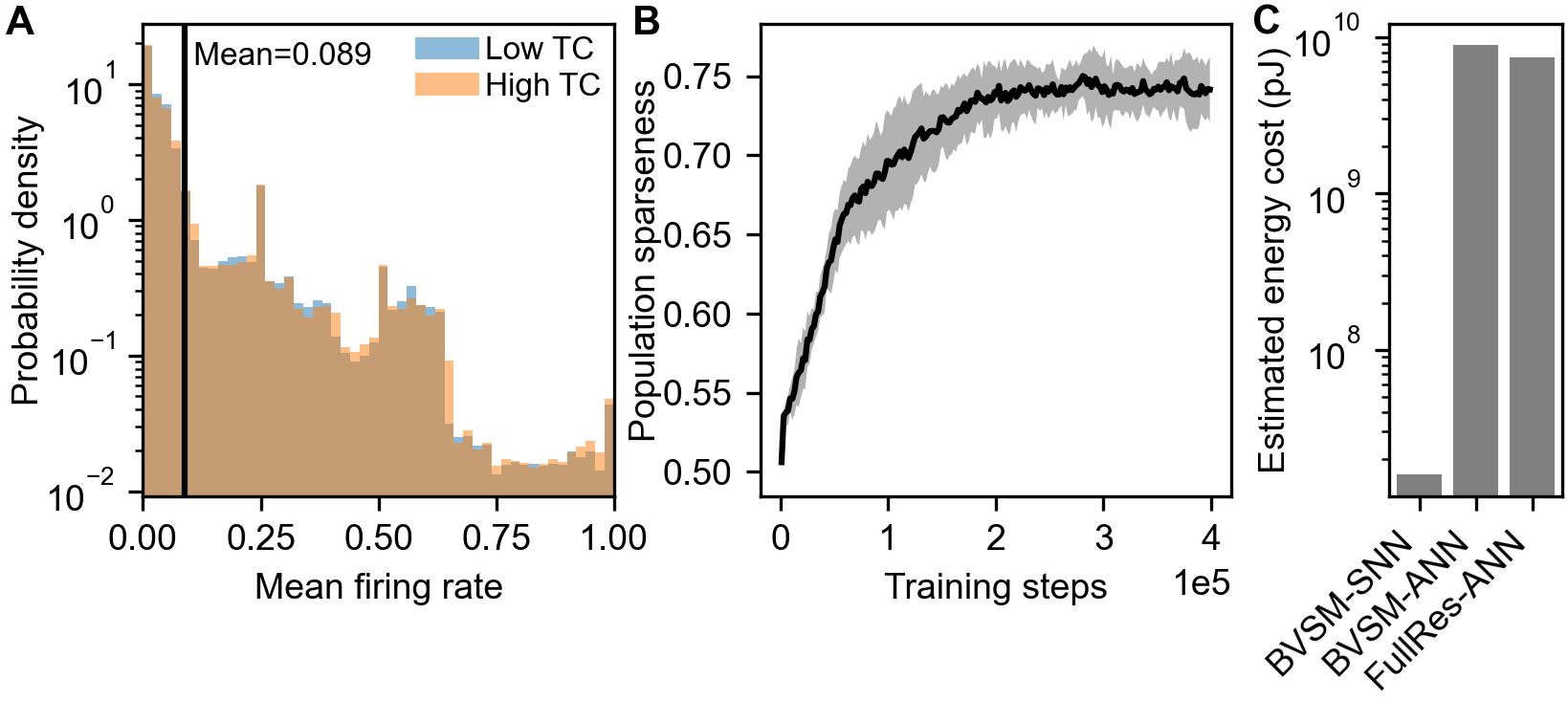}
    \caption{\textbf{A}: Distribution of neuron firing rate in BVSM. Firing rate is 1.0 if a neuron fires at every time step. TC: target contrast. \textbf{B}: The change in population sparseness of FEN during training. \textbf{C}: Estimated energy cost of BVSM (BVSM-SNN), BVSM based on ANN (BVSM-ANN), and conventional ANN that process full resolution images without making eye movements (FullRes-ANN). We assume BVSM-SNN and BVSM-ANN use on average 10 fixations to find the target.} 
    \label{fig:model_fr_sparse}
\end{figure}

Previous research has suggested that the human search strategy in this task is suboptimal due to the preference for short saccades \cite{Zhou2021}, which increases the overlapping area of central vision in consecutive fixations. To improve BVSM's search performance, the radius of the IOR reward function was increased to $2.5^\circ$ (similar to the size of FEN's effective visual field), and the saccade amplitude reward was changed to equation \ref{eq:sacamp_reward_linear}, which penalizes large saccades relatively weaker than equation \ref{eq:sacamp_reward_exp}. The entropy target parameter in SAC was also lowered to make the strategy less stochastic (Table~\ref{table:hyperparameter_list}). These modifications result in a stronger doughnut-shaped fixation strategy that is similar to the ELM searcher (Fig.~\ref{fig:model_elm_fixloc}), and the search speed is closer to the ELM searcher (Table~\ref{table:search_performance}, see methods in Appendix \ref{elm_model_appendix}).

\subsection{BVSM Has Low Estimated Energy Cost and High Robustness}

We estimate the energy consumption of BVSM, BVSM based on ANN, and conventional ANN that process full-resolution images without making eye movements (see methods in Appendix~\ref{energy-consumption-analysis-method}). BVSM's estimated energy consumption is two orders of magnitude lower than that of ANN counterparts, even if BVSM needs on average 10 fixations to find the target (Table~\ref{table:energy-estimation}, Fig.~\ref{fig:model_fr_sparse}C). This is because BVSM automatically develops sparse coding during training (Fig.~\ref{fig:model_fr_sparse}B, see methods in Appendix \ref{fr_sparseness_method}). Most neurons have a low firing rate (Fig.~\ref{fig:model_fr_sparse}A), and the average firing rate is only 0.089 (maximum possible firing rate is 1.0). The sparse activation in the network is similar to what was found in the animal brain \cite{Vinje2000, Weliky2003, Franco2007, Froudarakis2014, Yoshida2020}. The low saccade decision latency of only 4 time steps also increases the energy efficiency by preventing repeated processing of input images.

BVSM's search performance is robust to modest changes in the discount factor and the reward function (Table~\ref{table:hyperparameter_scan}). However, removing either of the two reward functions will cause a notable slowdown in search speed, and may also decrease the response accuracy (Table~\ref{table:hyperparameter_scan}). If we double the target's diameter, BVSM will search much faster and more accurately (Table~\ref{table:hyperparameter_scan}).

\section{Discussion} \label{chap:discussion}

\paragraph{Benefit of non-uniform resolution retina and eye movements}
Although non-uniform resolution retina and eye movements are common features of biological vision \cite{Clarke1976, Rowe1976, Dunlop1987, GonzalezSoriano1995, Bozzano2000, Tyrrell2013}, they are not widely used computer vision due to several possible reasons. Processing a single uniform-resolution image by a single forward computation is simpler, faster, and easier to train compared to selecting multiple fixations on the same image. Moreover, the use of multiple fixations prevents parallelization due to temporal dependency on successive fixations, and may offset the efficiency of non-uniform down-sampling.

However, the bio-inspired visual system has potential benefits that are yet to be explored. In real-life situations, robots do not process many single, unrelated images, but a continuous stream of visual input, so the saccade rate is very likely to be lower than the input frame rate, and making multiple fixations on the same image frame is unlikely to happen. Therefore, the benefit on computational efficiency of non-uniform sampling will be more significant. Besides, many computer vision tasks do not require the algorithm to simultaneously have large visual field, high visual resolution, low response latency, and low energy cost; but for robots that freely move in natural environments, these four features may likely be required, and the bio-inspired visual system may provide a good solution under this scenario. 

For the visual search task in this research, conventional network that processes full-resolution images fails to detect the target (Appendix \ref{full_resolution_model_appendix}), but non-uniform down-sampling can enlarge the relative size of the target, and make the detection easier. Another possible search method is to divide the image into several patches and process each patch until finding the target. The output of the BVSM's artificial retina is 1/8.4 of the size of the original image. If we divide the image into eight patches and examine each sequentially, the median and mean fixation number will always be 4.5. This sliding window strategy is quicker when the target size is very small and the background offers no hints of the target's location, so the peripheral vision is ineffective. However, violating either of these conditions can render the sliding window strategy inefficient.

For example, when the target's size doubles (to $0.6^\circ$), BVSM achieves 99.9\% search accuracy with a median fixation number of only two fixations, outpacing the sliding window strategy. In natural images, object locations are often statistically related, so the background can provide cues to target positions. As a result, we believe BVSM is crucial for efficient visual search, particularly when a large visual field and low computational cost are needed, and when low-resolution peripheral vision can offer information about the target's location.

\paragraph{Benefit of SNN} 
BVSM achieves low estimated energy cost mainly due to its sparse SNN and low saccade decision latency. Sparsity is important for avoiding spike congestion and decreasing hardware latency in real neuromorphic chips \cite{Peres2022}. BVSM automatically learns to use sparse coding without explicitly adding sparsity constraints during training. Future studies may investigate whether the model will also develop sparse coding on more complex natural images.

Compared to previous object detection SNNs with latencies of 100--3500 time steps \cite{Kim2020, Kim2021, Chakraborty2021, Kugele2021, Miquel2021, Shen2022aerospace}, BVSM makes saccade decisions in just 4 time steps and finds the target in 10--16 fixations on average. The low saccade decision latency is achieved by using population coding, which mimics the animal brain \cite{Averbeck2006}. Population coding decreases the SNN latency by only increasing the neurons in the last layer, so it is much more energy efficient than increasing the SNN latency which causes all neurons to fire more. It's a promising choice for future SNNs that encode continuous-valued output variables.

\paragraph{Human visual search strategy}
BVSM's search speed slows down with a more human-like search strategy. This provides another evidence that the human eye movement strategy is suboptimal in terms of search speed \cite{Zhou2021}. A possible cause of the suboptimality is the preference for small saccades \cite{Zhou2021}, which causes a large overlap of foveal regions between successive fixations. Similar saccade amplitude distribution has been observed in a wide range of tasks and stimulus images \citep{Bahill1975, Castelhano2009, Wang2011, Devillez2015}, so probably humans emphasis more on minimizing saccade cost (e.g. saccade duration and saccadic suppression) than using the strictly optimal policy. Additionally, the difficulty of the task may prompt individuals to make cautious small saccades to ensure thorough searching and avoid omissions.

Despite having better peripheral target visibility, human search accuracy is worse than the model. In detection task, subjects know the target location, so the visibility may be higher than that in visual search \citep{Najemnik2005, Zhou2021}. In detection task without target location cue, the response accuracy at fovea dropped to about 95\%, and the detection rate of target at periphery is still higher than the model (Fig.~\ref{fig:human_visibility_nocue}, Appendix \ref{vismap-experiment-appendix}). However, in visual search which involves multiple fixations, a lower foveal visibility may cause humans to find several suspected target locations and needs to compare between them, which slows down visual search and decreases the accuracy. This was what many subjects reported after finishing the experiment.

Measuring target visibility without a target location cue may seem like a more realistic reflection of the visual search experiment. However, in a sequence of eye movements, target visibility may increase at future fixation locations, even to the level comparable to the visibility measured with target location cue\citep{Gersch2004, Baldauf2008}. Therefore, the actual target visibility during visual search may fall between the visibility measured with and without target location cue, but it is difficult to measure since we can't accurately predict where humans will fixate in the future.

\paragraph{Limitations of BVSM}
BVSM is only tested on artificial images, requires GPU for training, and is not trained end-to-end, which hinders energy-saving during training. BVSM's retina is also simple compared to the human retina, and it only remembers the past fixation locations and lacks the diverse search strategies of humans, who may remember suspected target locations or scan the image for confirmation of an already found target. Future models could address these limitations by incorporating event camera input, maintaining more information in memory, and including fixational eye movement to improve coding efficiency and discrimination of fine details \citep{Rucci2007, Segal2015}. A world model and top-down attention \citep{Gazzaley2012, Mittal2020} could also be added to BVSM to improve its generalizability for diverse vision tasks. For real-life applications, we should also be careful about potential adversarial attacks that draw BVSM's attention to incorrect locations.

\paragraph{Conclusion}
We propose BVSM, the first bio-inspired model integrating non-uniform resolution retina, efficient eye movement strategy and SNN. It excels in executing visual searches with ultra-low saccade decision latency and sparse activations, which lay the foundation for integration with neuromorphic chips to achieve high energy efficiency. The comparison with human eye movement behavior provides another piece of evidence that the human search strategy is suboptimal on certain tasks, and it also reveals the limitations of BVSM. Future research should focus on enhancing the scalability and versatility of the model to handle a variety of visual tasks using eye movements in natural images.

\bibliography{reference}
\bibliographystyle{unsrtnat}

\newpage
\appendix

\renewcommand\thefigure{S\arabic{figure}}  
\setcounter{figure}{0}
\renewcommand\thetable{S\arabic{table}}
\setcounter{table}{0}
\renewcommand\thealgorithm{S\arabic{algorithm}}
\setcounter{algorithm}{0}
\renewcommand{\theequation}{S\arabic{equation}}
\setcounter{equation}{0}

\section{Human Experiment Details}\label{human-experiment-details}

\subsection{Human Subjects}\label{human-subjects}

29 human subjects (14 females, aged 19-30 years) participated in the study for monetary compensation. All subjects were tested to have normal or corrected-to-normal vision, and reported no neurological or psychological disease history. This study was approved by the Ethics Committee of our institution, and all subjects provided written informed consent.

\subsection{Apparatus}\label{apparatus}

We used a 24.5-inch BenQ ZOWIE XL2540 LCD monitor (resolution 1920×1080, physical size 54.3744×30.2616 cm, refresh rate 240 Hz) to present the stimuli. Stimuli were generated by custom MATLAB (MathWorks) scripts and presented using MATLAB and the Psychophysics Toolbox \citesupp{SuppKleinerMBrainardD2007, SuppPelli1997, SuppPrins2018}. Subjects were seated 70 cm away from the screen, and their heads were fixed by a forehead and chin rest. Binocular eye movements were recorded at 2,000 Hz with a TRACKPixx3 eye-tracker (Vpixx Technologies, Canada). Before each experiment, subjects first performed a 13-point calibration until the average test-retest measurement error of both eyes fell below 0.5$^\circ$. The eye-tracker was re-calibrated whenever the eye-tracking accuracy failed to meet the experiment requirements (see below).

\subsection{Target Contrast Selection Experiment} \label{supp_detection_exp_procedure}

This experiment used a two-interval-force-choice (2IFC) task to measure the relationship between foveal target visibility (\(d'\) in signal detection theory \citesupp{SuppMacmillan2005}) and target contrast, and selected a target contrast for each individual that has the same foveal visibility. 15 subjects participated in the experiment. The details of the procedure were reported in \citesupp{SuppZhou2021}.

In each trial, subjects fixated at image center and viewed two stimulus images for 250 ms each. One image was pure background noise, and the other was background noise with a target embedded at image center. Subjects had to select the image with the target. Within each trial, we used the eye-tracker to continuously monitor subject's fixation location. If the subject fixated more than 1$^\circ$ away from the image center, the trial would be aborted. We tested 4-5 levels of target contrast (ranging from 0.03 to 0.12), and each contrast level had 200 valid trials. In each trial, the target contrast is randomly selected from these levels. Visibility is calculated as \citesupp{SuppMacmillan2005}:
\begin{align} \label{eq:visibility}
d'=\frac{z(p(\text{Hit}))-z(1-p(\text{CR}))}{\sqrt{2}},
\end{align}
where $p(\text{Hit})$ is the hit rate, $p(\text{CR})$ is the correct rejection rate, and $z(x)$ is the inverse standard normal cumulative distribution function. The hit (or correct rejection) rate was defined as the number of trials that target appeared in the first (or second) interval, and the subject made a correct response, divided by the total number of trials that target appeared in the first (or second) interval. 

For each subject, we selected the target contrast that had a foveal visibility of 3.0 and used it for all following experiments. This was done by first fitting a Weibull function between the hit and correct rejection rate and target contrast, and then use equation~\ref{eq:visibility} to find the target contrast that makes $d'=3.0$. The Weibull function is fitted by maximum likelihood estimation implemented in the Palamedes Toolbox \citesupp{SuppPrins2018}. Therefore, each subject had a different target contrast (ranged 0.11-0.136).

\subsection{Target Visibility Map Experiment}\label{vismap-experiment-appendix}

This experiment measured the target visibility as a function of retinal location and fixation duration. The procedure was the same as the first task, except that the target contrast was fixed at the selected level, and the target was presented at different locations for different durations. The details of the procedure were reported in \citesupp{SuppZhou2021}.

The experiment had three versions. Four subjects participated in the first version, four subjects participated in the second version (one participated in both the first and second versions), and the rest eight subjects participated in the third version. 

In the first version, the subjects were required to fixate at the image center and identify which of the two stimulus intervals contained a target at a cued location. The experiment contained five sessions. In each session, the target appeared in one of the four cardinal directions relative to the image center, or appeared at the image center. We chose 3--4 different target locations (except when the target was at the image center) and 4--5 levels of stimulus duration in each session. When the target appeared at the image center, the stimulus duration was chosen from 4 to 200 ms, and each level had 100 trials. When the target appeared at peripheral locations, the stimulus duration was chosen from 50 to 700 ms, and each combination of stimulus duration and target location had 50 trials. There were about 5000 trials in total. The tested target locations were determined for each subject by a pilot experiment.

The second version was the same as the first version, except that the target location cue was not shown before each trial, and the target location and stimulus duration were randomly chosen from all possible combinations in each trial. There were about 5000 trials in total.

The third version was the same as the first version, except that there was only one level of stimulus presentation time (250 ms). It was used to measure the visibility map efficiently and give subjects some training to prepare for the visual search experiment. 

Subjects should maintain the fixation at the image center in each trial. We used the eye-tracker to continuously monitor subject's fixation location. If the subject fixated more than 1$^\circ$ away from the image center, the trial would be aborted. Aborted trials were not counted in the number of trials that the subject has finished.

\subsection{Target Visibility Data Analysis Method}

We use the visual information accumulation model proposed by \citetsupp{SuppZhou2021} to fit the experimentally measured target visibility data. The model describes a leaky integration of visual signals at each location over time within a fixation, and the relationship between target visibility $d'$ and location $(x,y)$ and fixation duration $T$ can be written as:
\begin{gather}
a(x,y) = p_{1} \cdot \exp\left( - p_{2} \cdot \sqrt{x^{2} + p_{5} \cdot y^{2}} \right), \label{eq:dp_a} \\ 
k(x,y) = p_{3} \cdot \exp\left( - p_{4} \cdot \sqrt{x^{2} + p_{5} \cdot y^{2}} \right), \label{eq:dp_k} \\
d'(x,y,T) = a(x,y)\sqrt{\frac{1 - \exp( - k(x,y)T)}{k(x,y)\left\lbrack 1 + \exp( - k(x,y)T) \right\rbrack}}, \label{eq:dp_vt_leakyint}
\end{gather}
where \(p_{1}\sim p_{5}\) are free parameters. The values of these parameters were fitted by minimizing the mean-squared error between the calculated and measured visibility data by the GlobalSearch algorithm in MATLAB's Global Optimization Toolbox.

\subsection{Visual Search Experiment}

In visual search task, subjects were told to search for the target as fast and accurately as possible, but there is no upper limit on search time. For the 15 subjects that participated in the detection task, the target contrast was the individually selected contrast (ranged 0.11-0.136, called low target contrast group). The rest 14 subjects did not participate in the detection task, and the target contrast was fixed to 0.15 (called high target contrast group). 

Each trial started by fixating within 0.8$^\circ$ away from the image center, and the target location was sampled uniformly within the region of background noise. Subjects were informed that the target could appear at anywhere within the image, even right at the initial fixation location. Subjects ended the trial by fixating on the target location and pressing a button. If the final fixation location is less than 1$^\circ$ away from the target, the trial is labeled as correct. Subjects first practiced for 100-150 trials, and performed 200-400 formal trials depending on their available time. We would re-calibrate the eye-tracker if subjects reported that they had correctly found the target, but the trial was wrong.

\subsection{Eye Tracking Data Analysis Method}

We analyzed eye-tracking data by the EYE-EEG extension \citesupp{SuppDimigen2011} of the EEGLAB toolbox \citesupp{SuppDelorme2004}. We used a velocity-based algorithm with adaptive threshold \citesupp{SuppNystrom2010} to classify eye movements into saccades, fixations, and blinks. Fixations separated by blinks were classified as two fixations. Glissades were merged into fixations. We used the minimum instantaneous eye movement velocity of both eyes in the eye movement event detection algorithm to suppress intermittent noise in the data of a single eye. 

Since each subject finished a different number of trials, when summarizing eye movement statistics, we first summarize the eye movement statistics within each subject, then average the statistics across different subjects. 

\section{Bayesian Optimal Visual Search Model}\label{elm_model_appendix}

We use the entropy-limit-minimization (ELM) searcher \citesupp{SuppNajemnik2009} as the optimal visual search model. The model's input is not the actual search image, but random signals generated according to human target visibility data. According to signal detection theory\citesupp{SuppMacmillan2005}, we assume that the visual signals obtained within one fixation from one location of the visual field obey a normal distribution. The presence of the search target at this location will shift the mean of the distribution, but will not change the variance. Visibility $d'$ is also defined as the difference between the mean divided by the common standard deviation. 

Theoretically, the ELM model should be able to fixate at any location in the image. In practice, to decrease the computational cost, the model could only fixate at 400 locations uniformly sampled inside the search image. We set the visual signal $W_{i,L_F}$ obtained at location $i$ ($i=1 \cdots n, n=400$) during fixation $F$ at location $L_F$ obeys the normal distribution $N(\pm 0.5, 1/d'^2_{i,L_F})$, where the mean is positive only when the target is at location $i$. $d'_{i,L_F}$ is calculated by equation \ref{eq:dp_vt_leakyint}, and the parameter $T$ in equation \ref{eq:dp_vt_leakyint} is fixed to 250 ms (close to the average fixation duration). Therefore, the visual signals gathered from all 400 locations is a vector $\textbf{W}_{L_F} = \left(W_{1,L_F}, \cdots, W_{n,L_F} \right)$. The posterior probability of target location given the visual signals gathered from all fixations can be calculated as:
\begin{align}\label{eq:p(i|W1-WF)}
P_{i,F} &= P\left( i\mid \mathbf{W}_{L_{1}},\cdots\mathbf{W}_{L_{F}} \right)  \notag \\
&= \frac{P\left( \mathbf{W}_{L_{1}},\cdots\mathbf{W}_{L_{F}} \mid i \right)p(i)}{\sum_{j = 1}^{n}\left\lbrack P\left( \mathbf{W}_{L_{1}},\cdots\mathbf{W}_{L_{F}}\mid j \right)p(j) \right\rbrack} \notag \\
&= \frac{e^{\sum_{f = 1}^{F}{W_{i,L_{f}}d_{i,L_{f}}^{\prime 2}}} \cdot p(i)}{\sum_{j = 1}^{n}\left\lbrack e^{\sum_{f = 1}^{F}{W_{j,L_{f}}d_{j,L_{f}}^{\prime 2}}} \cdot p(j) \right\rbrack} \notag \\
& = \frac{P_{i,F-1} \cdot e^{W_{i,L_{F}}d_{i,L_{F}}^{\prime 2}}}{\sum_{j = 1}^{n}\left( P_{j,F-1} \cdot e^{W_{j,L_{F}}d_{j,L_{F}}^{\prime 2}} \right)},
\end{align}
where $p(i)=1/400$ is the prior probability of target location.

The ELM searcher chooses the location that maximizes the expected information gained from the next fixation as the next fixation location. This can be expressed as:
\begin{gather}
H_{L_F} = - \sum_{i = 1}^{n}\left\lbrack P_{i,F} \cdot \ln P_{i,F} \right\rbrack, \notag \\
L_{F+1} = \underset{L_{F + 1}}{\text{argmax}}\left\{ H_{L_{F}} - E\left\lbrack H_{L_{F + 1}} \right\rbrack \right\}, \label{eq:ELM_rule}
\end{gather}
where \(E\lbrack \cdot \rbrack\) represents expectation with respect to the visual signal obtained in the next fixation. $E\left\lbrack H_{L_{F + 1}} \right\rbrack$ is an unknown variable that should be estimated, and $H_{L_F}$ is a known variable that can be calculated from the posterior probability of target location. \citetsupp{SuppNajemnik2009} derive a simple way to estimate $E\left\lbrack H_{L_{F + 1}} \right\rbrack$, and found that the next fixation location can be selected as:
\begin{align} \label{eq:ELM_rule_final}
L_{F+1} = \underset{L_{F + 1}}{\text{argmax}}\left\{ \frac{1}{2}\sum_{i = 1}^{n}\left( P_{i,F} \cdot d_{i,L_{F+1}}^{\prime 2} \right) \right\}.
\end{align}

When the posterior probability of target being at current fixation location exceeds a threshold \(\theta_{T}\), the ELM model sets the current fixation location as the responded target location. \(\theta_{T}\)is a constant value determined by matching the model's search accuracy to BVSM's search accuracy. In addition, the maximum number of fixations per trial is 200, and if this number is reached, the model will set the last fixation location as the responded target location. If the responded target location matches the actual target location, the trial is considered to be a correct trial.

\section{Bio-inspired Visual Search Model}\label{visual-search-model-details}

\subsection{Artificial Retina}\label{artificial-retina-appendix}

We modified the original Foveal Cartesian Geometry \citesupp{SuppMartinez2006} to make the transition of sampling density between central and peripheral vision to be more smooth. We define the radius (half of the width in pixels) \(r\) of a series of sampling rings (indexed by \(i\) starting from 1) as:
\begin{align} \label{eq:cfg}
r(i) = \left\{ \begin{matrix}
i,\ \ & i \leq i_{\text{fovea}} \\
a^{i + b} + c,\ \ & i > i_{\text{fovea}} \\
\end{matrix} \right.\ 
\end{align}
where \emph{a, b, c} are parameters, and \(i_{\text{fovea}}\) is the radius (in pixels) of fovea in the output image. The three parameters in equation~\ref{eq:cfg} should satisfy three constraints: 1. The two functions should intersect at \(i_{\text{fovea}}\); 2. \(r(i)\) should be smooth at \(i_{\text{fovea}}\) (first-order derivative is equal); 3. The radius of the outermost ring should be a predefined value \(r_{\max}\). This means that:
\begin{align} \label{eq:cfg_constraints}
\left\{ \begin{matrix}
a^{i_{\text{fovea}} + b} + c = i_{\text{fovea}} \\
a^{i_{\text{fovea}} + b} \cdot \ln a = 1 \\
a^{i_{\max} + b} + c = r_{\max} \\
\end{matrix} \right.\ 
\end{align}
In practice, the systems of non-linear equations can be solved by a modification of the Levenberg-Marquardt algorithm implemented in SciPy (\emph{scipy.optimize.root(method='lm')}) \citesupp{SuppMore1980}. We set the size of output image to be 224×224 with a fovea size of \(16 \times 16\ \)pixels (\(i_{\max} = 112,\ i_{\text{fovea}} = 8\)). To make sure the entire search image is always within the visual field (as is the case for humans), we set \(r_{\max}\) to be twice the radius of the search image.

On each ring, we uniformly sample \((8*i - 4)\) number of pixels. This means that pixels within the fovea are directly copied to the output, and pixels outside of the fovea are sampled progressively sparser. The sampled pixels are stacked together to form a square image as the output. We implemented the above retinal transformation by using the Kornia package \citesupp{SuppEriba2019Kornia}.

\subsection{Feature Extraction Network}\label{cnn-structure-appendix}

The feature extraction network contains 7 convolution blocks. The first convolution block uses 7×7 convolution, and subsequent blocks use 3×3 convolution. The number of output channels of the \emph{i\textsuperscript{th}} block is 16×\emph{i}. We use replicate padding in convolution, and down-sampling is achieved by setting stride $=2$. The output of the last convolution block is flattened into a 448-dimensional vector and fed into the three fully connected heads. Each fully connected head contains one layer of neurons. 

To convert FEN to SNN, we directly replace the QCFS function with the scaled integrate-and-fire (IF) neuron\citesupp{Suppbu2021optimal}: 
\begin{align} \label{eq:IF_neuron}
\left\{ \begin{aligned} 
&\widetilde{\mathbfit{v}}_{l,t} = \mathbfit{v}_{l,t-1} + \frac{\mathbf{W}_l \mathbfit{s}_{l-1,t} + \mathbfit{b}_l}{\lambda} \\
& \widetilde{\mathbfit{s}}_{l,t} = H\left( \widetilde{\mathbfit{v}}_{l,t} - \theta \right)\\
&\mathbfit{s}_{l,t} = \lambda \widetilde{\mathbfit{s}}_{l,t} \\
&\mathbfit{v}_{l,t} = \widetilde{\mathbfit{v}}_{l,t} - \widetilde{\mathbfit{s}}_{l,t}\theta_l \\
\end{aligned} \right.\ 
\end{align}
where $l$ is the layer index, $t$ is the SNN inference time index, $\mathbfit{v}_{l,t}$ is the membrane potential of IF neurons in layer $l$ at time $t$ ($\mathbfit{v}_{l,0} = 0.5$), $\mathbfit{s}_{l,t}$ is the scaled output spikes from layer $l$ at time $t$, \(\theta = 1\) is the firing threshold, \(H\left( \cdot \right)\) is the Heaviside step function, $\mathbf{W}_l$ and $\mathbfit{b}_l$ are the connection weights and bias from layer $l-1$ to layer $l$, and $\lambda$ is the parameter $\lambda$ in QCFS function. When a neuron fires, we reset its membrane potential by subtracting \(\theta\) to prevent information loss \citesupp{SuppRueckauer2017}.

We directly use the pixel values (repeated for $T=4$ time steps) as the network's input. Given input images with batch size \(B\), the output of the converted SNN has dimension \(\left( T \times B \times 5 \right)\).

\subsection{Recurrent Spiking Network}\label{spiking-rnn-appendix}

Denote SNN inference time as $T$ and batch size as $B$. At each fixation $l$, the network takes the predicted fixation location $\mathbf{x}_{l}$ ($T\times B\times 2$) and the RNN hidden state $\mathbf{h}_{l}$ ($B\times 64$) as input. Within each fixation, at the first SNN time step, the network linearly transforms $\mathbf{x}_{l}[1,:,:]$ and $\mathbf{h}_{l}$. The sum of the results is then passed into IF neurons and produces output spikes $\mathbf{s}_{l}[1,:,:]$. At subsequent time steps $t$, the network only linearly transforms $\mathbf{x}_{l}[t,:,:]$ and produce the output spikes $\mathbf{s}_{l}[t,:,:]$. The output spikes from the last time step become the RNN hidden state $\mathbf{h}_{l+1}$ for the next fixation. This computation process is also shown in Algorithm \ref{alg:SRNN}. The IF neurons are reset to the initial membrane potential (0.5) at the start of each fixation. In practice, the last linear transform in FEN and the initial linear transform in RNN can be merged together, allowing spikes to straightforwardly transmit from FEN to RNN.

\begin{algorithm}[htb!]
   \caption{Recurrent spiking network}
   \label{alg:SRNN}
\begin{algorithmic}
   \STATE {\bfseries Input:} \(\mathbf{x}\) with dimension
        \(\left( T,B,L,2 \right)\), initial RNN hidden state
        \(\mathbf{h}_{0}\) with dimension \((B,64)\) (filled with zeros). $T$ is the SNN inference time. $B$ is batch size. $L$ is the number of fixations.
    \STATE Initialize output spikes \(\mathbf{s}\) with dimension \((T,B,L,64)\). 
    \STATE Load linear transform weight matrix $\mathbf{W}_{xr}$, $\mathbf{W}_{rr}$ and bias $\mathbf{b}_{x}$, $\mathbf{b}_{r}$.

   \FOR{$l=1$ {\bfseries to} $L$}
        \STATE Reset IF neuron membrane potential to 0.5.
        \FOR{$t=1$ {\bfseries to} $T$}
            \STATE  $\mathbf{y} = \mathbf{x}[t,:,l,:] \cdot \mathbf{W}_{xr} + \mathbf{b}_{x}$
            \IF{t == 1}
                \STATE $\mathbf{y} = \mathbf{y} + \mathbf{h}_{l-1} \cdot \mathbf{W}_{rr} + \mathbf{b}_{r}$
            \ENDIF
            \STATE  $\mathbf{s}[t,:,l,:] = \text{IF}(\mathbf{y})$ according to equation~\ref{eq:IF_neuron}.
            \STATE Output spikes $\mathbf{s}[t,:,l,:]$.
        \ENDFOR
        \STATE  $\mathbf{h}_{l} = \mathbf{s}[T,:,l,:]$
   \ENDFOR
   \STATE {\bfseries Output:} RNN hidden state $\mathbf{h}_{L}$

\end{algorithmic}
\end{algorithm}

\subsection{Actor and Critic Networks} \label{actor-critic-appendix}
\begin{figure}[htb!]
    \centering
    \includegraphics[width=1.0\textwidth]{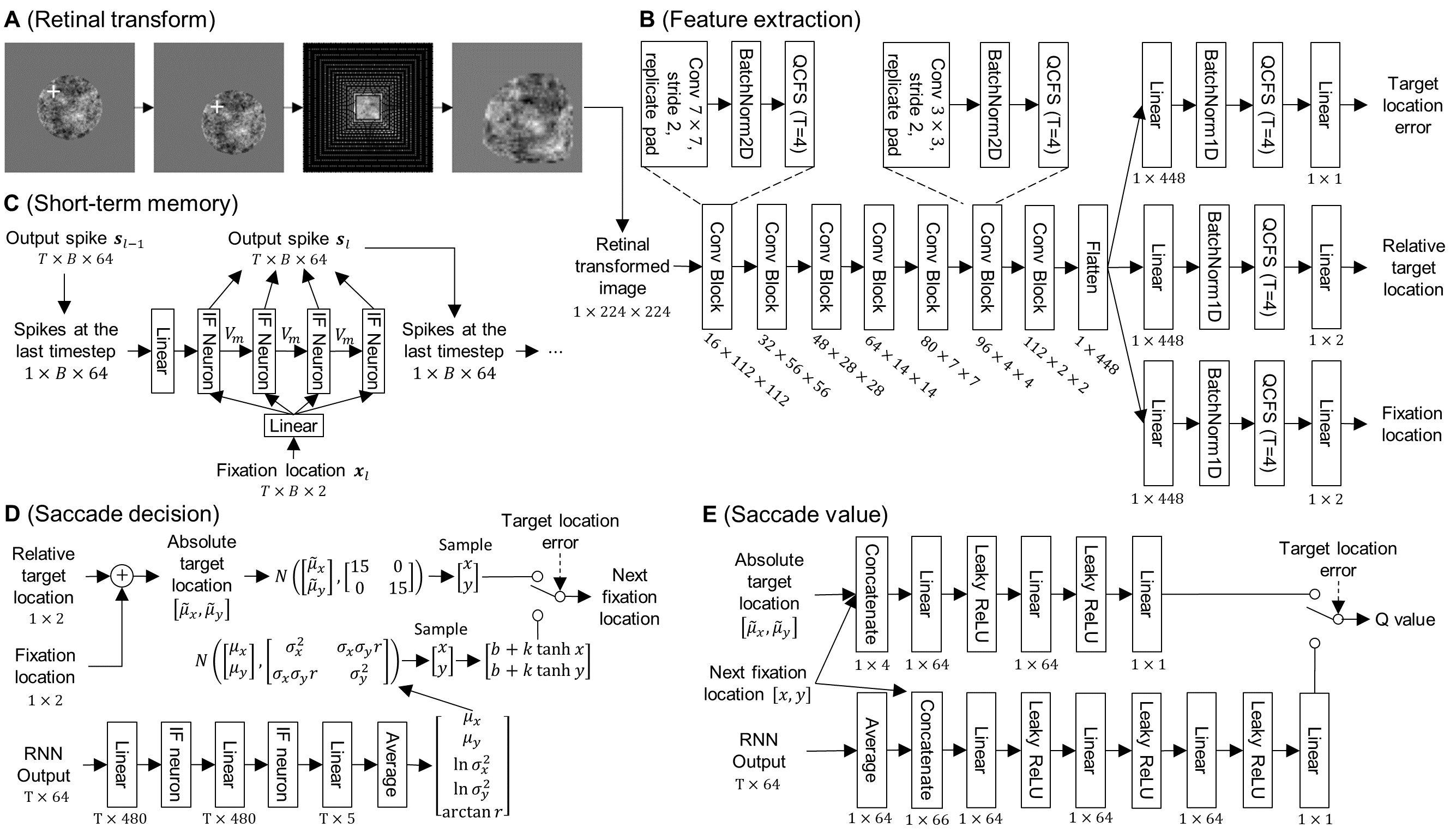}
    \caption{
        Detailed structure of BVSM. \textbf{A}. Retinal transform procedure. First: original search image. White cross indicates fixation location. Second: translated image with fixation location at the center. Third: non-uniform sampling. Only the shown pixels are sampled. Fourth: the sampled pixels are stacked together to form the retinal transformed image. \textbf{B}. Feature extraction network structure. The dashed lines show the internal components of convolution block. Only the first convolution block uses $7\times 7$ convolution, and all subsequent blocks use $3\times 3$ convolution. QCFS: quantization clip-floor-shift function. The number under each block indicates the dimension of the output when batch size $B=1$. \textbf{C}. Spiking RNN network structure with dimension of input and output variables. $T=4$: SNN inference time. $B$: batch size. $l$: The $l^{th}$ fixation. $V_m$: membrane potential of the IF neuron. This panel shows the computation of spiking RNN within a single fixation unrolled in SNN inference time. The four IF neuron blocks are the same group of neurons at each SNN time step. \textbf{D}. Actor network structure. It is an SNN of IF neurons. $N(M, \Sigma)$ represents Gaussian distribution with mean \(M\) and covariance matrix \(\Sigma\). Parameters $k$ and $b$ transform the sampled action from normalized unit [-1,1] to pixels [0,650]. The number under each block indicates the dimension of the output when batch size $B=1$. \textbf{E}. Critic network structure. It is an ANN with leaky-ReLU (negative slope = 0.1) activation function. The number under each block indicates the dimension of the output when batch size $B=1$. Average means averaging over time \emph{T}. In D and E the arrow with dash line indicates that the switch is controlled by the target location error.
        }
    \label{fig:model_structure}
\end{figure}

The actor network makes saccade decisions. If the estimated target location error from FEN is lower than a threshold (0.58$^\circ$, or 25 pixels), the next fixation location will be sampled from a 2-dimensional Gaussian distribution centered at the predicted target location with a small symmetrical covariance matrix of 15 pixel\textsuperscript{2}; otherwise the actor will pass the RNN output into two layers of 480-dimension IF neurons (equation \ref{eq:IF_neuron}), and linearly transform the result into a 5-dimensional vector (population coding), which encodes the mean and covariance of the next fixation location's distribution (Fig.~\ref{fig:model_structure}D). The fixation location output by the actor can be anywhere within the square region enclosing the circular search image.

The critic network measures saccade value (Fig.~\ref{fig:model_structure}E). If the estimated target location error is low, it will concatenate the next fixation location with the predicted absolute target location and output a Q-value. Otherwise it will concatenate the next fixation location with RNN output, and calculate another Q-value. The critic network is not a spiking network, as it is not used in inference after training. Therefore, the output from the RNN needs to be averaged over the first dimension (representing SNN time steps) before concatenation.

\subsection{Detection Task by FEN}\label{cnn-detection-appendix}

We use a 2IFC task similar to human experiment to measure the FEN's ability to correctly locate the target as a function of target eccentricity. In each trial the network is presented with two images, one contains the background alone, and the other contains the background and a target. We select the image whose estimated target location error is below the detection threshold (25 pixels), and the predicted target location is less than 1$^\circ$ away from the true target location. We use the true target location in decision rule because the target location is cued in human experiment. If the decision rule is not matched for both images, then we will randomly select an image. A trial is correct if the selected image contains the target.

We measure the correct response rate of 6 target contrasts (0.11 to 0.135, step 0.005), 49 target eccentricities (0.0 to 0.98 of the image radius, step 0.02), and 8 target angular positions (0$^\circ$ to 315$^\circ$, step 45$^\circ$). Each combination of conditions contains 400 trials. The relationship between proportion correct and target distance is fitted by the Weibull function (with inverted x-axis, because Weibull function is monotonically increasing) by maximum likelihood estimation implemented in the Palamedes toolbox \citesupp{SuppPrins2018}.

\subsection{BVSM Training and Evaluation Method}\label{model-training-appendix}

In the first training stage, we train the FEN with QCFS function as activation function by supervised learning. The network is trained for 174,000 optimization steps. In the second training stage, we convert the trained FEN to spiking FEN, and then fine-tune the spiking network for 35,000 optimization steps by supervised learning. Fine-tuning is done by STBP implemented in spikingjelly \citesupp{SuppSpikingJelly}. All parameters remain trainable. We use the Atan function to approximate the gradient of a spike \citesupp{SuppFang2021}.

The training data in the first two stages are random samples of retinal transformed images generated at runtime during training. We first generate random samples of full-resolution background noise images. For each image, we randomly set a fixation location uniformly within the circular region. We then randomly set one target location per image whose distance to fixation location obeys exponential distribution with a mean of 4$^\circ$. The target location has to be within the circular noise image, otherwise we will resample another target location. This ensures that enough training samples contain the target near the foveal region and helps the network to converge more reliably.

In the third training stage, we fix the FEN, and use SAC \citesupp{SuppHaarnoja2018} algorithm to train the RNN, actor, and critic networks to do visual search. The SAC agent contains one FEN, one RNN with one target network, one actor, and two critics with two target networks. We use a replay buffer of 50,000 trials (each trial can have a variable number of fixations). The state vector stored in replay buffer is the FEN output in each fixation. Optimization started after the initial 333 trials, and is performed once per fixation. In each optimization, we randomly sample 32 trials (with variable number of fixations) from the replay buffer, update critic and RNN by temporal difference learning, update actor alone by gradient ascent through the critic output, and finally update the SAC temperature parameter according to \citesupp{SuppHaarnoja2018}. The gradients in RNN and actor are calculated by STBP and surrogate gradient (Atan function). The gradients of RNN and critic are clipped to a maximum norm of 1.0. The number of training trials is 50,000. A list of hyperparameters can be found in Table~\ref{table:hyperparameter_list}.

During training, target contrast is randomly chosen within {[}0.11, 0.15{]}. Target and initial fixation locations are randomly chosen within the circular search image. All three stages of training were performed on a computer with Intel i9 9900K CPU and NVIDIA GeForce RTX 2080Super GPU. The first and second stages required less than 5 hours, and the third stage required about 20 hours. The code is implemented in PyTorch \citesupp{SuppPyTorch2019}. 

During evaluation, target location is randomly chosen within the circular search image, but the initial fixation location is randomly chosen within 0.35$^\circ$ away from the image center. This is similar to human experiment where a visual search trial always started from fixating at the image center. The target contrast is chosen within {[}0.11, 0.136{]} (low target contrast) or fixed to 0.15 (high target contrast) during evaluation, corresponding to the target contrast range in human experiment. The model is evaluated for 10,000 visual search trials for each target contrast level.

\subsection{Firing Rate and Sparseness Analysis Method}\label{fr_sparseness_method}

We record the activities of every neuron in BVSM during evaluation. The firing rate of each neuron is calculated as the number of time steps that the neuron fires divided by the total number of time steps in 10,000 visual search trials. The total number of time steps is the total number of fixations multiplied by SNN latency (4 time steps) in each fixation. Therefore, the firing rate of each neuron falls in the region between 0 and 1.

BVSM has 359,504 neurons in total, and 99.72\% of them are in the FEN. To calculate the change of population sparseness during training, we calculate the change of population sparseness during the first training stage of FEN. Note that in the first training stage, the FEN is not an SNN, but the QCFS activation function \citesupp{Suppbu2021optimal} outputs the firing rate (scaled by $\lambda$ in QCFS function) of each neuron. Therefore, we can calculate the population sparseness \citesupp{SuppWeliky2003, SuppYu2014, SuppYoshida2020} by the unscaled QCFS output:
\begin{gather}
r_i = \text{clip}\left( \frac{1}{T} \text{floor}\left(\frac{xT}{\lambda} + 0.5 \right),0,1 \right), \\
S = \frac{1}{1-\frac{1}{N}} \left[1 - \frac{\left(\sum_{i=1}^{N}\frac{r_i}{N} \right)^2}{\sum_{i=1}^{N}\frac{r_{i}^2}{N}} \right],
\end{gather}
where $i$ represents neuron index and $N$ is the total number of neurons. To obtain the final stable population sparseness during the FEN training process, we increase the number of training steps from 175,000 to 400,000. But when running visual search, we still use the parameters saved at 175,000 training steps.

\subsection{Energy Consumption Analysis Method}\label{energy-consumption-analysis-method}

We compare the energy consumption of BVSM with two different models. One is BVSM based on conventional artificial neural networks, the other is a conventional image classification network that directly processes original full-resolution images. BVSM based on ANN can be constructed by replacing IF neurons in BVSM to leaky-ReLU activation function, and its search performance is similar to BVSM. The conventional full-resolution ANN has a similar convolution backbone to the feature extraction network in BVSM (Fig.~\ref{fig:fullres_net_structure}), but is deeper since the input size is larger.

We use the technical data of Intel Stratix 10 TX FPGA chip \citesupp{SuppIntel2022} and ROLLS neuromorphic chip \citesupp{SuppIndiveri2015} to estimate the energy consumption of BVSM and the other two networks. The FPGA consumes 12.5 pJ per FLOP, while the neuromorphic chip consumes 0.077 pJ per synaptic operation (SOP) and 3.7 pJ per spike. The same data were used in prior works \citesupp{SuppHu2021, Suppbu2021optimal} to estimate the energy consumption of ANN and SNN. 

For SNN, one SOP is defined as the transmission of one spike from a single neuron to another single neuron. We count the average number of spikes that each neuron fires within each fixation from 10000 visual search trials. The number of post-synaptic neurons of each neuron can also be calculated from the network structure. We do not account for the energy consumed by the floating point computation in the first convolution layer, but the energy consumed by the spikes generated in the first layer is included. The reason for this is that we lack energy consumption data for neuromorphic chips under a constant floating point input. Additionally, some neuromorphic chips, such as Intel Loihi, quantize the floating point into low-precision fixed-point representations, and the energy spent on this process is difficult to estimate accurately. To ensure a fair comparison, we also disregarded the floating point operation in the first convolution layer of the ANN.

For ANN, one FLOP is defined as the elementary calculation between two scalars. The FLOP of a convolution layer is $2 \times C_{in} \times K \times K \times C_{out} \times H_{out} \times W_{out}$, where $C$ is channels, $K$ is the convolution kernel size, and $H, W$ are the size of the output. The FLOP of a linear layer is $2 \times W_{in} \times W_{out}$, where $W_{in}$ and $W_{out}$ are input and output dimensions. Batch normalization can be merged into convolution and linear layers, so they do not add additional FLOPs. We also assume that the ANN uses simple activation functions (such as ReLU or Leaky-ReLU), so the FLOP of activation function is simply the number of output neurons of each layer.

\section{Full-resolution Image Classification Model}\label{full_resolution_model_appendix}

\begin{figure}[htb!]
    \centering
    \includegraphics[width=1.0\textwidth]{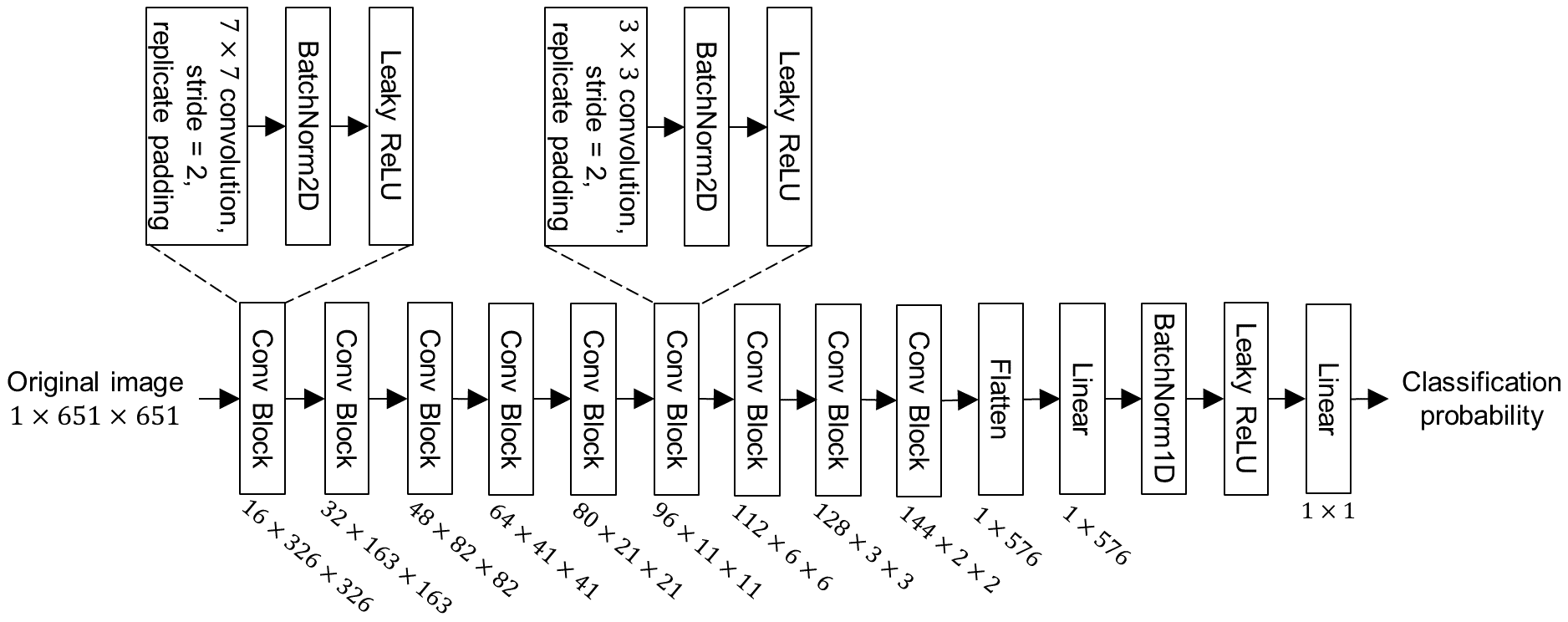}
    \caption{Structure of the conventional ANN that processes full resolution image and classifies whether the image contains a target. The dashed lines show the internal components of convolution block. Only the first convolution block uses $7\times 7$ convolution, and all subsequent blocks use $3\times 3$ convolution. The leaky-ReLU function has slope of 0.1 in the negative part. The number under each block indicates the dimension of the output when batch size $B=1$.} 
    \label{fig:fullres_net_structure}
\end{figure}

To test whether conventional ANN that directly processes full-resolution images can detect the target, we build a convolution network that has a similar backbone structure to the FEN (Fig.~\ref{fig:fullres_net_structure}). The network performs a binary classification task, which is to determine whether the input image contains a target or not. This is easier than the visual search task which needs the network to output the exact target location. The network outputs the classification probability by a single forward computation without making eye movements. 

\begin{figure}[htb!]
    \centering
    \includegraphics[width=0.95\textwidth]{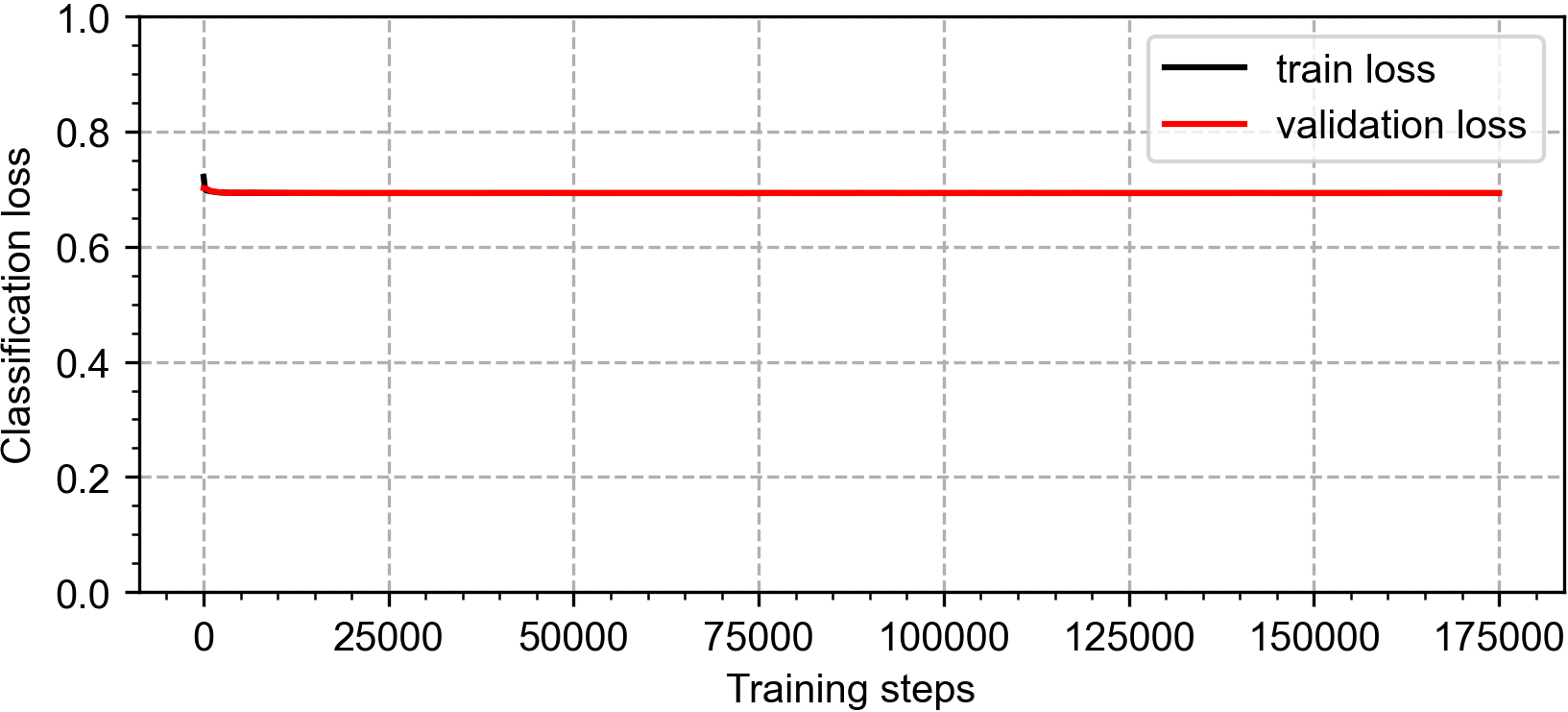}
    \caption{The training and validation loss of conventional ANN that processes full-resolution images and performs binary image classification task. The networks are trained using five different random seeds. The loss values are first smoothed by a Gaussian kernel with standard deviation of 100 steps, then average across the five repetitions. The shaded area represents the standard deviation of the smoothed loss.}
    \label{fig:cnn_fullres_training_process}
\end{figure}

We generate random samples of the search images, and train the ANN by supervised learning. The loss function is binary cross-entropy, the optimizer is AdamW without weight decay, the learning rate is 1e-3, and the training batch size is 64. We train the network for 175,000 steps (same as the training step in the first training stage of FEN) under 5 different random seeds, but the loss value cannot decrease (Fig.~\ref{fig:cnn_fullres_training_process}), and the network's correct response rate is only at chance level (50\%). Therefore, conventional ANN that directly processes full-resolution images cannot properly detect the small search target. 

\clearpage
\newpage

\section{Supplementary Tables}

\begin{table}[htb!]
\caption{Comparison of our task and model design to previous research. Target's size, background image's size, and target's relative size are only shown for visual search tasks. Variable fixation number means that the model can automatically stop the trial without hard coding it to make decision after a predefined number of fixations. Training method refers to the method to train eye movement policy. EM: expectation maximization. RL: reinforcement learning. SVHN: The Street View House Numbers dataset. IOR: inhibition of return. (1): This is the size of the smallest bounding box enclosing all targets. The size is calculated based on the authors' data preprocessing method. (2): A single bounding box may contain 2-5 targets. (3): The target size, image size, and target relative size are calculated from the dataset.}
\begin{center}
\begin{tiny}
\setlength{\tabcolsep}{2pt}
\begin{tabulary}{1.0\textwidth}{CCCCCCCCCC}
\toprule
\textbf{Source} & \textbf{Task} & \textbf{Target size} & \textbf{Background size} & \textbf{Target relative size} & \textbf{Artificial fovea} & \textbf{Memory} & \textbf{SNN} & \textbf{Number of fixations} & \textbf{Training method} \\ \midrule
\citesupp{SuppRanzato2014} & MNIST search \& recognition & 28x28 & 48x48 & 34.0\% & Multi-resolution patches & Mean of predictions at each fixation & No & Fixed & Similar to EM algorithm \\ \midrule
\citesupp{SuppMnih2014}    & MNIST search \& recognition & 28x28 & 60x60 & 21.8\% & Multi-resolution patches & RNN & No & Fixed & RL \\ \midrule
\multirow{2}{*}{\citesupp{SuppBa2015}} & MNIST search \& recognition (multiple targets) & 28x28 & 100x100 & 7.84\% & Multi-resolution patches & RNN & No & Fixed per target & RL \\ \cline{2-10}  
                                & SVHN recognition & 49x49\textsuperscript{(1)} & 54x54 & 16.47\%-41.17\%\textsuperscript{(2)} & Multi-resolution patches & RNN & No & Fixed per target & RL \\ \midrule
\citesupp{SuppDauce2020}  & Noisy MNIST search \& recognition & 28x28 & 128x128 & 4.79\% & Log-Polar transform & No & No & Variable & Supervised \\ \midrule
\citesupp{SuppDabane2022} & Noisy MNIST search \& recognition & 28x28 & 128x128 & 4.79\% & Log-Polar transform & No & No & Fixed & Supervised \\ \midrule
\citesupp{SuppKumari2022} & MNIST search \& recognition & 28x28 & 60x60 & 21.78\% & Multi-resolution patches & RNN & No & Variable & RL \\ \midrule 
\citesupp{SuppAlexe2012} & Natural object search & (2-499)x(4-499)\textsuperscript{(3)} & (142-500) x (71-500)\textsuperscript{(3)} & Median: 9.71\%, mean: 20.82\%\textsuperscript{(3)} & Single patch & Exponential decay memory & No & Fixed & Supervised \\ \midrule
\citesupp{SuppAkbas2017} & Natural object search & (4-499)x(4-499)\textsuperscript{(3)} & (127-500) x (96-500)\textsuperscript{(3)} & Median: 6.51\%, mean: 16.67\%\textsuperscript{(3)} & Simplified Freeman-Simoncelli model & Bayesian prior, manual IOR & No & Variable & Maximum-a-posteriori \\ \midrule
\citesupp{Suppelsayed2019saccader} & Image classification & N/A & N/A & N/A & Single patch & Mean of predictions at each fixation, manual IOR & No & Fixed & RL \\ \midrule
\citesupp{Supprangrej2022consistenc} & Image classification & N/A & N/A & N/A & Single patch & Transformer & No & Training: fixed, Testing: variable & RL \\ \midrule
\citesupp{Supprangrej2023glitr} & Video classification & N/A & N/A & N/A & Single patch & Transformer & No & Fixed & Supervised \\ \midrule
Our model & Gabor target search inside noisy images & 12x12 & 651x651 & 0.034\% & Foveal Cartesian Geometry & Spiking RNN & Yes & Variable & RL \\ \bottomrule
\end{tabulary}
\end{tiny}
\end{center}
\label{table:compare_previous_work}
\end{table}

\begin{table}[htb!]
\caption{Comparison of SNN configurations in our model and previous works for object detection and RL. IF: integrate-and-fire neuron. LIF: leaky integrate-and-fire neuron. STDP: spike-timing-dependent plasticity. STBP: spatio-temporal back-propagation. (1): The inference time depends on the length of event-based input.}
\begin{center}
\begin{tiny}
\setlength{\tabcolsep}{2pt}
\begin{tabulary}{1.0\textwidth}{CCCCCCCC}
\toprule
\textbf{Source}  & \textbf{Task}  & \textbf{Image type} & \textbf{RL algorithm}  & \textbf{Neuron type}  & \textbf{Coding method} & \textbf{Training method}  & \textbf{Latency} \\ \midrule
\citesupp{SuppKim2020}  & Object detection  & Natural image & N/A & Signed IF  & Rate coding  & Conversion                   & $>$3500      \\ \midrule
\citesupp{SuppKim2021}  & Object detection  & Natural image & N/A & IF         & Rate coding  & Bayesian optimization        & $>$500  \\ \midrule
\citesupp{SuppChakraborty2021} & Object detection & Natural image & N/A & LIF   & Rate coding & Conversion + STDP + STBP     & 300     \\ \midrule
\citesupp{SuppKugele2021}  & Object detection  & Neuromorphic image & N/A & LIF & Spike count & STBP       & 100-300\textsuperscript{(1)} \\ \midrule
\citesupp{SuppMiquel2021}  & Object detection  & Natural image      & N/A  & IF & Rate coding & Conversion & $>$1000 \\ \midrule
\citesupp{SuppShen2022aerospace} & Object detection & Aerospace image & N/A & Modified LIF & Rate coding & Conversion  & 64-128  \\ \midrule
\citesupp{SuppPatel2019}  & RL (Breakout)  & N/A  & Deep Q-Learning  & IF  & Rate coding & Conversion & 500  \\ \midrule
\citesupp{SuppTang2020} & RL (HalfCheetah-v3, Hopper-v3, Walker2d-v3, Ant-v3) & N/A & DDPG, TD3, SAC, PPO & LIF     & Population coding & STBP  & 5  \\ \midrule
\citesupp{SuppTan2021}  & RL (17 Atari games) & N/A & Deep Q-Learning & IF      & Rate coding   & Conversion & $>$100 \\ \midrule
\citesupp{SuppChowdhury2021} & RL (Cartpole, Pong) & N/A & Deep Q-Learning & LIF+custom non-spiking neuron & Membrane potential of non-spiking neuron & Conversion + STBP + iterative retraining & 1  \\ \midrule
\citesupp{SuppXu2022pop} & RL (Custom navigation task) & N/A & DDPG & Modified LIF & Population coding & STBP & 5 \\ \midrule
\citesupp{Suppchen2022deep} & RL (17 Atari games) & N/A & Deep Q-Learning & LIF+custom non-spiking neuron & Membrane potential of non-spiking neuron & STBP & 8 \\ \midrule
Out model        & Detection \& RL (visual search)  & Naturalistic noise image  & SAC & IF   & Population coding  & Conversion + STBP & 4       \\ \bottomrule

\end{tabulary}
\end{tiny}
\end{center}
\label{table:compare_snn_latency}
\end{table}

\begin{table}[htb!]
\caption{Hyperparameters for the results reported in the main text. IF neuron: integrate-and-fire neuron. FEN: feature extraction network. RNN: recurrent neural network. RL: reinforcement learning. SacAmp: saccade amplitude.}
\begin{center}
\begin{scriptsize}
\setlength{\tabcolsep}{2pt}
\begin{tabulary}{1.0\textwidth}{CCC}
\toprule
\textbf{Group}  & \textbf{Description}  & \textbf{Value}  \\ \cline{1-3}
\multirow{15}{*}{Visual search task}  & Screen distance  & 70 cm  \\ 
                                      & Screen size& 1920×1080 pixels, 54.3744×30.2616 cm  \\ 
                                      & Search image size & 15$^\circ$×15$^\circ$ (651×651 pixels), circular shape  \\ 
                                      & Search image contrast & 0.2  \\ 
                                      & Target size  & 0.3$^\circ$×0.3$^\circ$ (12×12 pixels)  \\ 
                                      & Target orientation  & 45$^\circ$   counterclockwise from the vertical  \\ 
                                      & Target spatial frequency  & 6 cycles/degree  \\ 
                                      & Target contrast  & Training: {[}0.11--0.15{]}. Testing: {[}0.11--0.136{]} or 0.15   \\ 
                                      & Target number per image  & 1   \\ 
                                      & Target location  & Uniformly random in search image   \\ 
                                      & Initial fixation location  & Training: uniformly random in image. \\
                                      & & Testing: uniformly random within 0.35$^\circ$ away from image center.  \\ 
                                      & Maximum fixation number  & 50 during training, 200 during testing   \\ 
                                      & Correct criterion  & Final fixation within 1$^\circ$ from target location   \\ \midrule
\multirow{3}{*}{Retinal transform}    & Method  & Modified Foveal Cartesian Geometry \\ 
                                      & Fovea radius  & 8 pixels   \\ 
                                      & Maximum peripheral radius  & 651 pixels  \\ 
                                      & Output size  & 224×224 pixels   \\ \midrule
\multirow{3}{*}{Reward function}      & Radius of IOR reward  & 2.5$^\circ$   \\ 
                                      & Memory capacity of IOR reward  & 8 previous fixations   \\ 
                                      & Slope of saccade amplitude reward  & -1/7.5$^\circ$   \\ \midrule
\multirow{5}{*}{IF neuron}    & Inference time  & 4 steps   \\ 
                              & Surrogate gradient function  & Atan()   \\ 
                              & Initial membrane potential  & 0.5   \\ 
                              & Firing threshold  & 1   \\ 
                              & Reset method after spike  & Reset by subtracting firing threshold   \\ \midrule
\multirow{6}{*}{FEN}  & Convolution kernel size  & 7×7 at first layer, 3×3 at subsequent layers  \\ 
                      & Convolution stride  & 2   \\ 
                      & Padding  & Replicate padding, padding size=$(\text{kernel size} - 1)/2$   \\ 
                      & Output channel number at block $i$  & 16×$i$   \\ 
                      & Activation function  & QCFS(Initial $\lambda$=8, $T$=4), $\lambda$ is trainable   \\ 
                      & Fully connected layer hidden dimension   & 1   layer, 448 neurons   \\ \midrule
RNN                   & Number of IF spiking neurons  & 64    \\ \midrule
\multirow{3}{*}{Actor network}   & Number   of hidden layers  & 2    \\ 
                                 & Hidden layer dimension  & 480 IF neurons   \\ 
                                 & Target detection threshold  & 25 pixels   \\ \midrule
\multirow{3}{*}{Critic network}  & Number of hidden layers  & 2 and 3 for the two heads  \\ 
                                 & Hidden layer dimension  & 64 for both heads   \\ 
                                 & Activation function  & Leaky-ReLU, negative slope = 0.1    \\ \midrule
\multirow{6}{*}{FEN training}  & Training batch size  & 64   \\ 
                               & Validation batch size  & 256    \\ 
                               & Training steps  & 174,000 steps for training, 35,000 steps for fine-tuning   \\ 
                               & Optimizer  & AdamW \citesupp{SuppLoshchilov2019} with no weight decay   \\ 
                               & Initial learning rate  & 1.00E-03   \\ 
                               & Distance between target and fixation locations & Exponential distribution with mean=4$^\circ$   \\ \midrule
\multirow{13}{*}{RL}  & Algorithm  & Soft Actor-Critic   \\ 
                      & Entropy target  & Hyperparameter group 1: -1; hyperparameter group 2: -2   \\ 
                      & Initial temperature $\alpha$  & 1    \\ 
                      & Discount factor$\gamma$    & 0.95   \\ 
                      & Replay buffer size  & 50,000 trials   \\ 
                      & Batch size in optimization   & 32 trials   \\ 
                      & Optimizer  & AdamW with no weight decay   \\ 
                      & Initial learning rate  & 1e-4 for actor and $\alpha$, 1e-3 for critic and RNN    \\ 
                      & Gradient clipping   & Maximum norm of 1.0 for critic and RNN   \\ 
                      & Initial optimization time   & After 333 trials    \\ 
                      & Optimization frequency  & Once per fixation   \\ 
                      & Target network update speed  & Polyak update, $\tau$=0.005   \\ 
                      & Training trials  & 50,000    \\ \midrule
\multirow{4}{*}{Reward functions}   & IOR reward radius  & Hyperparameter group 1: $0.5^\circ$, Hyperparameter group 2: $2.5^\circ$   \\ 
                            & IOR reward memory capacity  & 8 most recent fixations   \\ 
                            & Saccade amplitude reward (exponential) & Hyperparameter group 1: $-0.5 + 0.5 \times \left( \exp{\left(-\text{SacAmp} / 2.5^\circ \right)} - 1 \right)$ \\
                            & Saccade amplitude reward (linear) & Hyperparameter group 2: $-\text{SacAmp}/7.5^\circ$ \\ \bottomrule
\end{tabulary}
\end{scriptsize}
\end{center}
\label{table:hyperparameter_list}
\end{table}

\begin{table}[htb!]
\caption{Comparison of the estimated energy consumption between BVSM and conventional ANN models.}
\begin{center}
\begin{footnotesize}
\setlength{\tabcolsep}{4pt}
\begin{tabulary}{1.0\textwidth}{CCCCC}
\toprule
\textbf{Model} & \textbf{Number of FLOP} & \textbf{Number of spikes} & \textbf{Number of synaptic operations} & \textbf{Estimated energy cost} \\ \midrule
BVSM (SNN)         & -- & $1.28\times 10^5$/fixation & $1.45\times 10^7$/fixation & $1.59\times 10^6$ pJ/fixation  \\
BVSM (ANN)        & $7.10\times 10^7$/fixation & -- & -- & $8.88\times 10^8$ pJ/fixation \\ 
Full-resolution ANN   & $5.95\times 10^8$/image & -- & -- & $7.44\times 10^9$ pJ/image \\ 
\bottomrule
\end{tabulary}
\end{footnotesize}
\end{center}
\label{table:energy-estimation}
\end{table}

\begin{table}[htb!]
\caption{The search performance of BVSM under different hyperparameter settings. The model is trained with 5 different random seeds, and evaluated for 10000 visual search trials. The values of unchanged hyperparameters are from hyperparameter group 2. Saccade amplitude (SacAmp) reward (linear) refers to equation \ref{eq:sacamp_reward_linear}, SacAmp reward (exp) refers to equation \ref{eq:sacamp_reward_exp}. Median and mean fixation number (FixNum) are only calculated from correct trials.}
\begin{center}
\begin{footnotesize}
\setlength{\tabcolsep}{2pt}
\begin{tabulary}{1.0\textwidth}{CCCCCCCC}
\toprule
\textbf{Changed parameter} & \textbf{New value} & \textbf{Target contrast} & \textbf{\# of random seeds} & \textbf{Percent correct} & \textbf{Median FixNum} & \textbf{Mean FixNum} \\ \midrule
No change           & --    & 0.11--0.136 & 5   & 98.2\%  & 7.0  & 10.8  \\
Target diameter     & 0.6°  & 0.11--0.136 & 5   & 99.9\%  & 2.0  & 4.2   \\
Discount $\gamma$   & 0.99  & 0.11--0.136 & 5   & 98.0\%  & 7.4  & 10.9  \\ 
IOR memory capacity & 1     & 0.11--0.136 & 5   & 98.0\%  & 7.0  & 11.5  \\
IOR memory capacity & 2     & 0.11--0.136 & 5   & 98.1\%  & 7.0  & 11.4  \\ 
IOR memory capacity & 4     & 0.11--0.136 & 5   & 98.1\%  & 7.0  & 11.2  \\ 
IOR radius          & 1.5°    & 0.11--0.136 & 5   & 97.9\%  & 8.2  & 12.5  \\ 
Slope of SacAmp reward (linear)   & -1/4.5°    & 0.11--0.136 & 5   & 98.0\%  & 7.8   & 11.3  \\ 
Reward function  & Only IOR reward             & 0.11--0.136  & 5  & 97.1\%  & 9.8   & 17.4  \\ 
Reward function  & Only SacAmp reward (linear) & 0.11--0.136  & 5  & 83.3\%  & 21.6  & 44.2  \\ 
Reward function  & Only SacAmp reward (exp)    & 0.11--0.136  & 5  & 82.0\%  & 21.8   & 45.4  \\ 
\bottomrule
\end{tabulary}
\end{footnotesize}
\end{center}
\label{table:hyperparameter_scan}
\end{table}

\clearpage
\newpage

\section{Supplementary Figures}

\begin{figure}[htb!]
    \centering
    \includegraphics[width=0.8\textwidth]{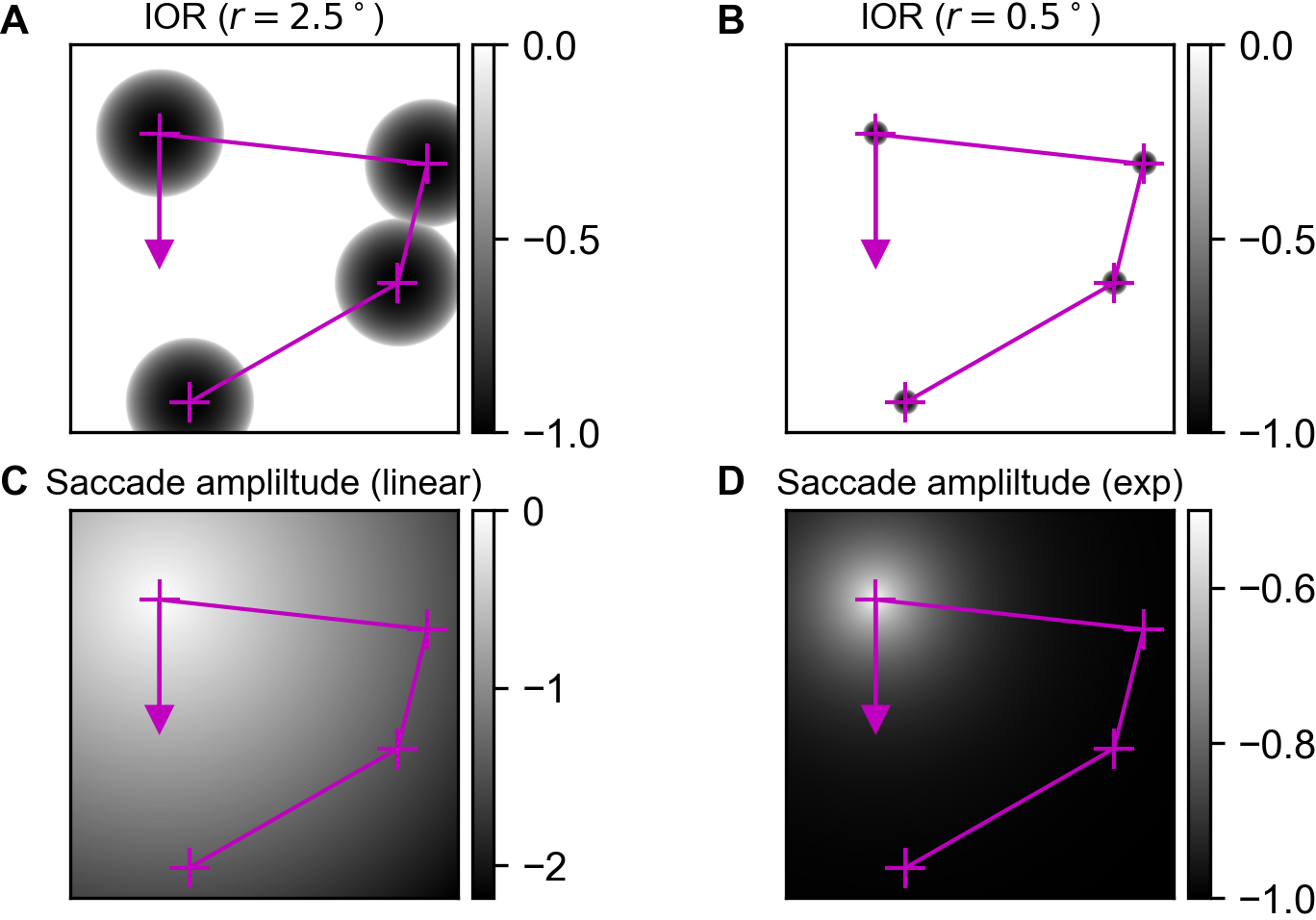}
    \caption{Demonstrations of reward functions. \textbf{A}, \textbf{B}: Inhibition of return reward. The semi-circle function (equation~\ref{eq:ior_reward}) has a radius of 2.5$^\circ$ and 0.5$^\circ$. \textbf{C}, \textbf{D}: Linear saccade amplitude reward (equation~\ref{eq:sacamp_reward_linear}) and exponential saccade amplitude reward (equation~\ref{eq:sacamp_reward_exp}). In all panels, the four magenta crosses show four most recent fixation locations, and the arrow shows an undetermined future fixation location. Each pixel in the image shows the reward values if the next fixation falls at this location.}
    \label{fig:reward_functions}
\end{figure}

\begin{figure}[htb!]
    \centering
    \includegraphics[width=0.95\textwidth]{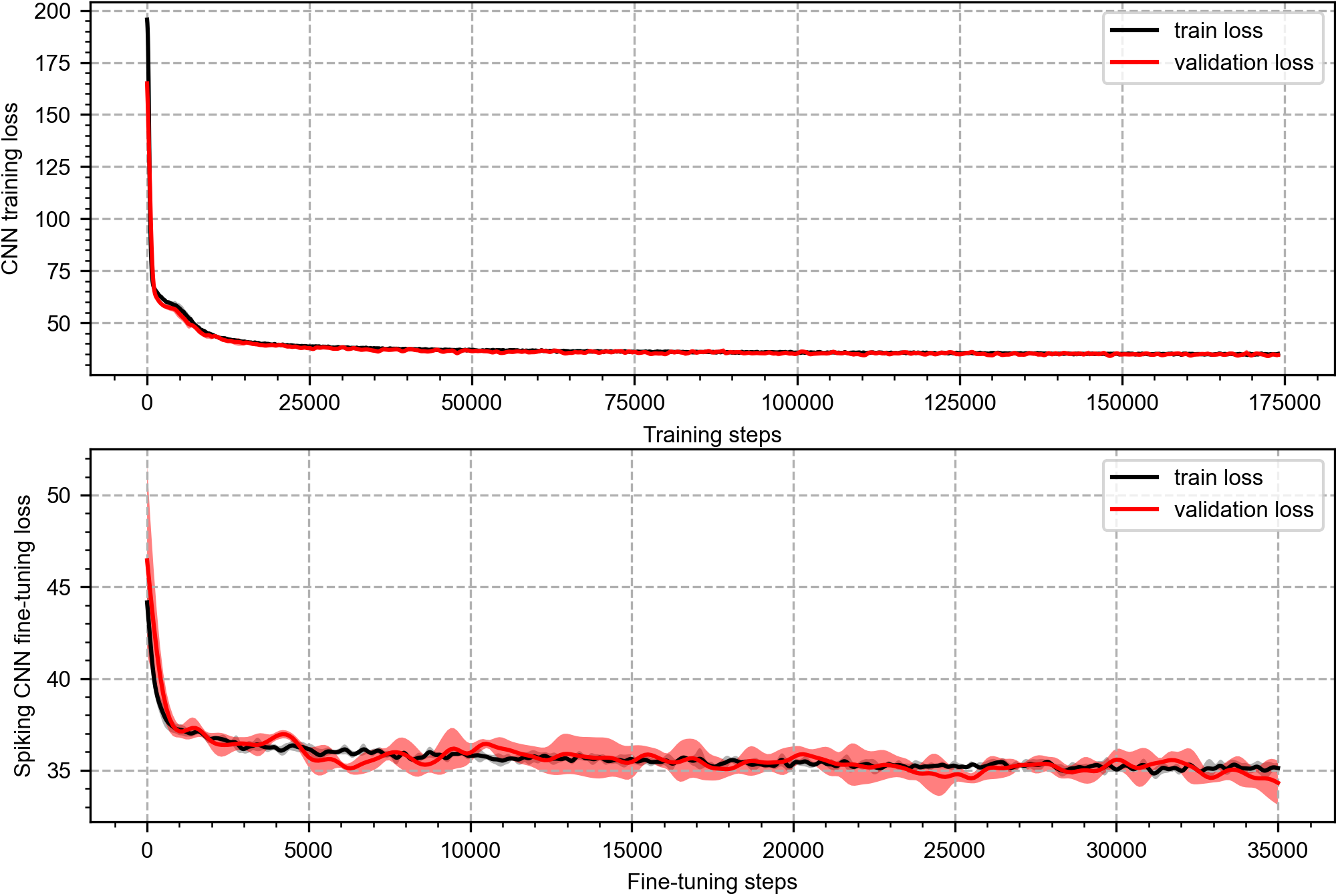}
    \caption{Up: Feature extraction network (FEN) initial training/validation loss. Low: Spiking FEN fine-tuning training/validation loss. The networks are trained using five different random seeds. The loss values are first smoothed by a Gaussian kernel with standard deviation of 100 steps, then average across the five repetitions. The shaded area represents the standard deviation of the smoothed loss.}
    \label{fig:cnn_training_process}
\end{figure}

\begin{figure}[htb!]
    \centering
    \includegraphics[width=0.95\textwidth]{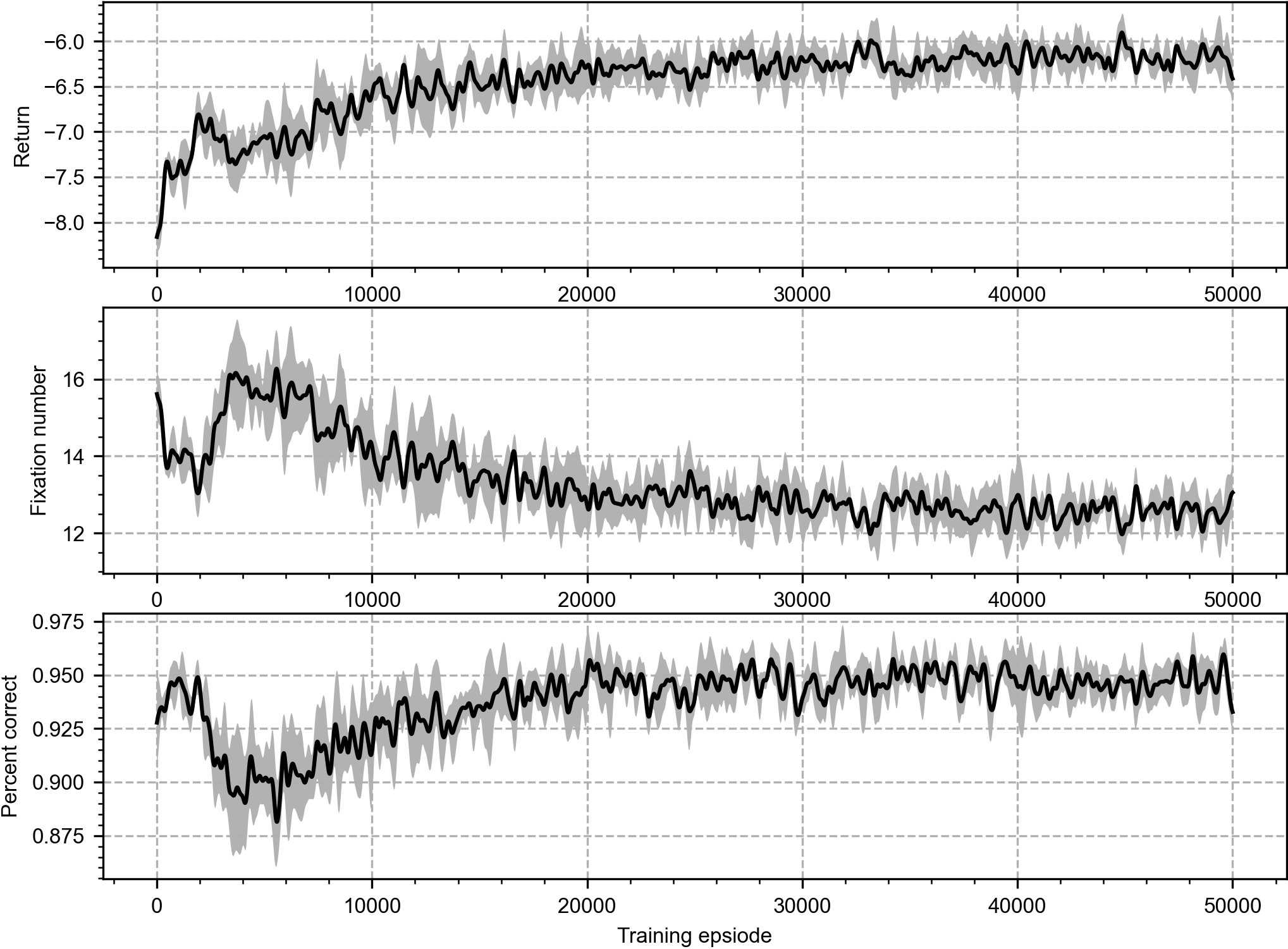}
    \caption{RL training process averaged across five different random seeds for hyperparameter group 1. The values are first smoothed by a Gaussian kernel with standard deviation of 100 steps, then averaged across the five repetitions. The shaded area represents the standard deviation of the smoothed values. Note that the fixation number is averaged across both correct and error trials, so it is higher than the mean and median fixation number of only correct trials reported in the main text. During training, the maximum fixation is 50, so the percent correct is lower than the result reported in the main text.}
    \label{fig:rl_training_process_hp1}
\end{figure}

\begin{figure}[htb!]
    \centering
    \includegraphics[width=0.95\textwidth]{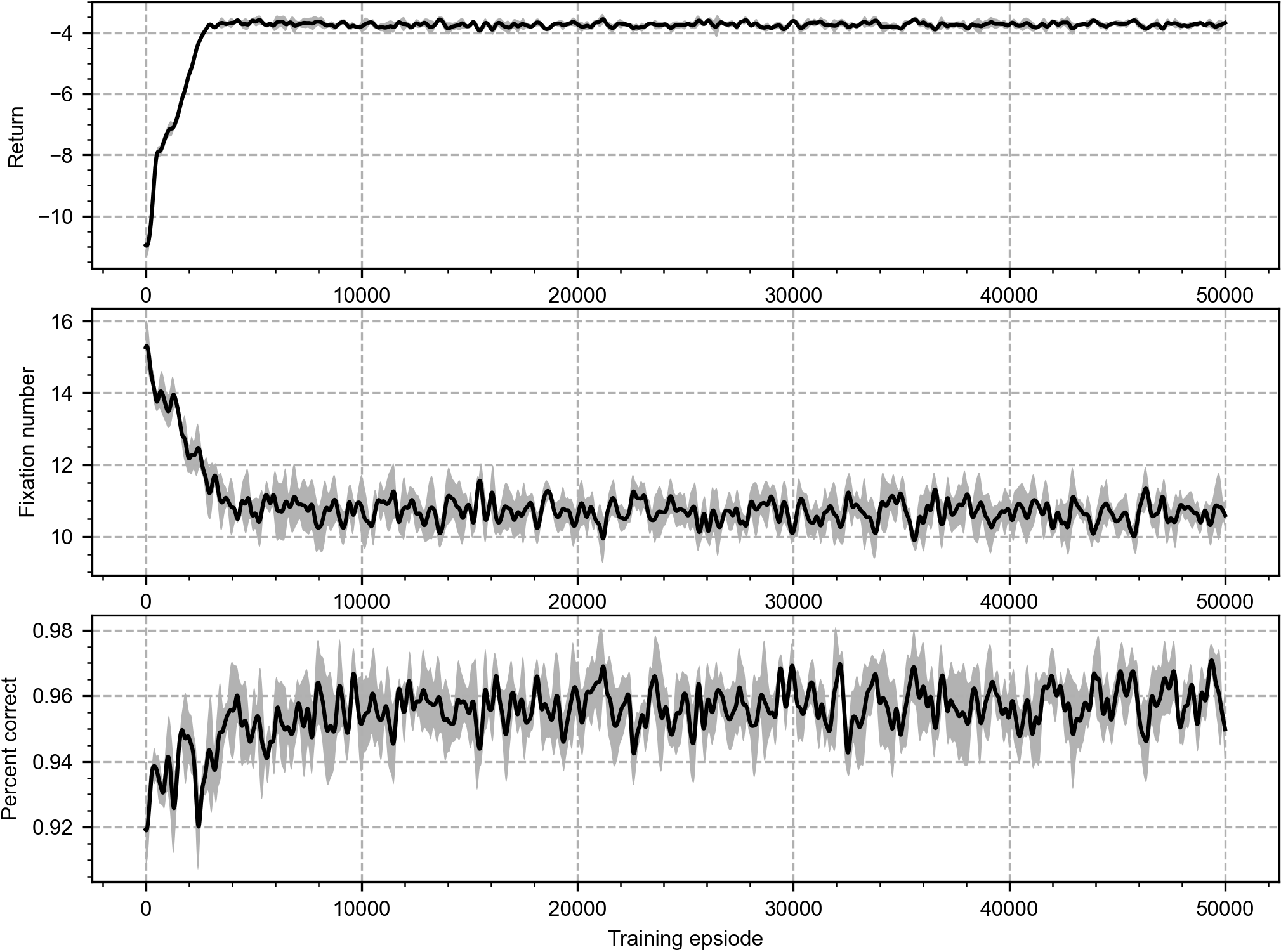}
    \caption{RL training process averaged across five different random seeds for hyperparameter group 2. The networks are trained using five different random seeds. The loss values are first smoothed by a Gaussian kernel with standard deviation of 100 steps, then average across the five repetitions. The shaded area represents the standard deviation of the smoothed loss.}
    \label{fig:rl_training_process_hp2}
\end{figure}

\begin{figure}[htb!]
    \centering
    \includegraphics[width=0.95\textwidth]{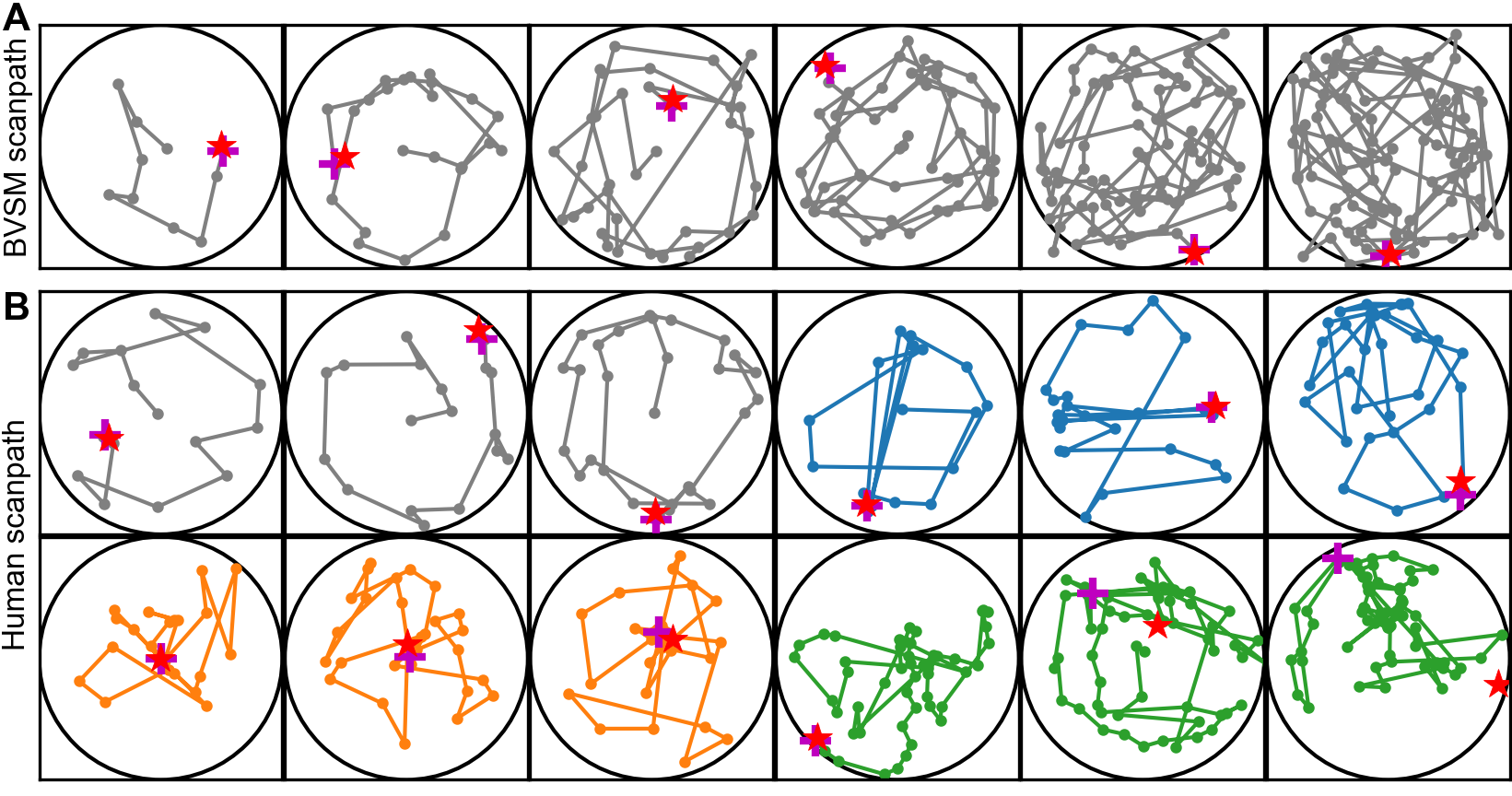}
    \caption{BVSM and human scanpath comparison. \textbf{A}: Randomly selected BVSM scanpath sorted by fixation number. Magenta cross is the final fixation location, red star is the target location. BVSM is trained by hyperparameter group 1 (HP 1). \textbf{B}: Example human scanpath grouped by different search strategies. Gray: circular scanpath. Blue: repeated fixation on suspected target locations. Yellow: confirmation scanpath when the target is close to the initial fixation location. Blue: scanpath with many short saccades.} 
    \label{fig:model_human_scanpath}
\end{figure}

\begin{figure}[htb!]
    \centering
    \includegraphics[width=0.6\textwidth]{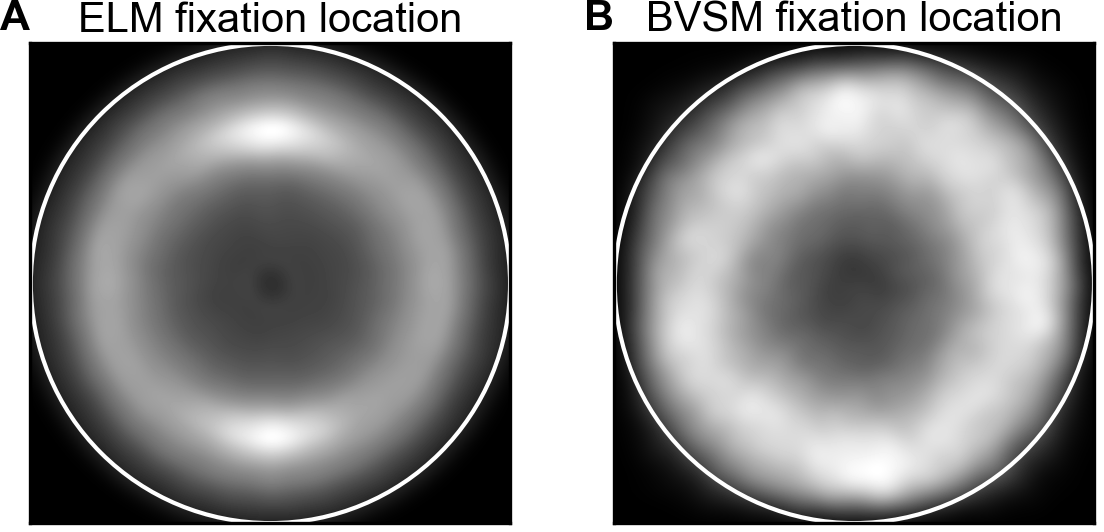}
    \caption{Distribution of fixation location of BVSM (\textbf{A}) and the optiaml ELM searcher (\textbf{A}). BVSM model is trained with hyperparameter group 2 (HP 2)} 
    \label{fig:model_elm_fixloc}
\end{figure}

\begin{figure}[htb!]
    \centering
    \includegraphics[width=0.3\textwidth]{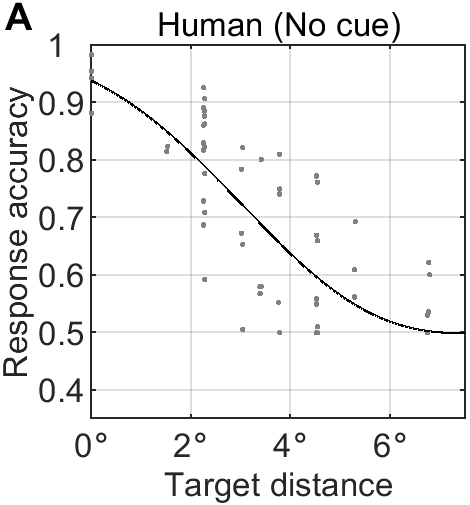}
    \caption{Human target visibility measured without target location cue. Each gray dot represents a combination of subject index, target eccentricity, and target angular position. The black curve represents the fitted Weibull function (with inverted x-axis).}
    \label{fig:human_visibility_nocue}
\end{figure}

\clearpage
\newpage

\bibliographysupp{reference_supp}
\bibliographystylesupp{unsrtnat}

\end{document}